\documentclass{article}

\usepackage{microtype}
\usepackage{graphicx}
\usepackage{subfigure}
\usepackage{booktabs} 
\usepackage{arydshln}
\usepackage{dsfont}
\usepackage{listings}
\usepackage{minitoc}
\usepackage{longtable}
\usepackage[toc,page,header]{appendix}
\usepackage{lmodern}

\usepackage{hyperref}

\usepackage[x11names, rgb, html, dvipsnames]{xcolor}

\usepackage{solarized-light}

\usepackage[accepted]{icml2024}

\usepackage{amsmath}
\usepackage{amssymb}
\usepackage{mathtools}
\usepackage{amsthm}
\usepackage{fancyvrb} 
\usepackage{verbatim} 

\usepackage[capitalize,noabbrev]{cleveref}

\theoremstyle{plain}

\theoremstyle{definition}

\theoremstyle{remark}

\newcommand{\myparagraph}[1]{\noindent\textbf{#1}}

\usepackage[textsize=tiny]{todonotes}

\icmltitlerunning{MARL Meets Leaf Sequencing in RT}

\begin{document}
\twocolumn[
\icmltitle{Multi-Agent Reinforcement Learning Meets Leaf Sequencing in Radiotherapy}




\begin{icmlauthorlist}
\icmlauthor{Riqiang Gao}{dti}
\icmlauthor{Florin C. Ghesu}{dtide}
\icmlauthor{Simon Arberet}{dti}
\icmlauthor{Shahab Basiri}{varian}
\icmlauthor{Esa Kuusela}{varian}
\icmlauthor{Martin Kraus}{dtide}
\icmlauthor{Dorin Comaniciu}{dti}
\icmlauthor{Ali Kamen}{dti}

\end{icmlauthorlist}

\icmlaffiliation{dti}{Digital Technology and Innovation, Siemens Healthineers, Princeton NJ, USA}
\icmlaffiliation{dtide}{Digital Technology and Innovation, Siemens Healthineers, Erlangen, Germany}
\icmlaffiliation{varian}{Varian Medical Systems, Siemens Healthineers, Helsinki, Finland}

\icmlcorrespondingauthor{Riqiang Gao}{riqiang.gao@siemens-healthineers.com}

\icmlkeywords{Machine Learning, ICML}

\vskip 0.3in
]
\RecustomVerbatimCommand{\VerbatimInput}{VerbatimInput}%
{fontsize=\footnotesize,
 %
 framesep=2em, 
 rulecolor=\color{Gray},
 %
 %
 commandchars=\|\(\), 
 commentchar=*        
}


\printAffiliationsAndNotice{}  
  
\doparttoc
\faketableofcontents

\begin{abstract}
In contemporary radiotherapy planning (RTP), a key module leaf sequencing is predominantly addressed by optimization-based approaches. In this paper, we propose a novel deep reinforcement learning (DRL) model termed as \textit{Reinforced Leaf Sequencer} (RLS) in a multi-agent framework for leaf sequencing. The RLS model offers improvements to time-consuming iterative optimization steps via large-scale training and can control movement patterns through the design of reward mechanisms.
We have conducted experiments on four datasets with four metrics and compared our model with a leading optimization sequencer. Our findings reveal that the proposed RLS model can achieve  reduced fluence reconstruction errors, and potential faster convergence when integrated in an optimization planner. Additionally, RLS has shown promising results in a full artificial intelligence RTP pipeline. We hope this pioneer multi-agent RL leaf sequencer can foster future research on machine learning for RTP.

\end{abstract}

\section{Introduction}
\label{introduction}

Radiotherapy (RT) stands as an essential cornerstone in the realm of cancer treatment, recommended for approximately half of cancer patients \cite{huynh2020artificial}.  Despite technological advances, a substantial part of RT still depends on labor-intensive and time-consuming planning pipelines from a diverse healthcare team \cite{elmore2019global}.  

\begin{figure}
    \centering
    \includegraphics[width=0.43\textwidth]{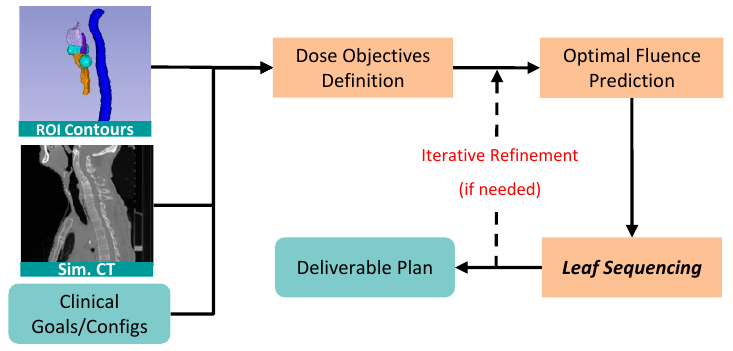}
    \caption{Illustration of a typical RTP process. Three common components are shown in the orange boxes. We focus on \textit{leaf sequencing} in this work. The term ``optimization" in this paper refers to a series of methods that are not machine learning. 
    }
    \label{fig:PTP}
    \vspace{-.2in}
\end{figure}

Radiotherapy Planning (RTP), originates from 1890s, refers to the process to plan the appropriate external beam RT treatment for patients with cancer. Modern RTP involves tailoring personalized treatment plans to effectively target tumors through Planning Target Volume (PTV) while safeguarding healthy tissues, as indicated by Organs at Risk (OARs). In practical RT, the RTP should also consider that the plan can be delivered in a reasonable time. Current RTPs are dominant by optimization pipelines in practical platforms such as Varian \cite{varian2023treat} and Elekta \cite{ElektaMCO}.  
 Though different treatment types are existing, e.g., intensity-modulated radiation therapy (IMRT) \cite{mundt2005intensity} and volumetric-modulated arc therapy (VMAT) \cite{otto2008volumetric}, their optimization pipelines share similar key components:  \textit{dose objectives definition},  \textit{optimal fluence prediction}, and \textit{leaf sequencing} to get feasible machine parameters,  as shown in Figure \ref{fig:PTP}\footnote{In-depth RTP introduction (background, terms, visuals, etc.) is in Appendix \ref{background}. Some leaf sequencing optimizations for VMAT skip fluence prediction. Here we mainly discuss the branch with fluence prediction in the pipeline, such as RapidArc$^\text{TM}$ (Varian).}. 
\textit{Optimal fluence prediction} estimates a 2D  intensity map of the radiation beam for each beam angle allowing for customizable dose delivery to meet the objectives, however, the practicality of the fluence may be limited as it could potentially violate certain physical rules necessary for achievability. \textit{Leaf sequencing} is to approximate optimal fluences with feasible machine movements. Depending on the implementation details of algorithms, the fluence map optimization and leaf sequencing can be performed sequentially and iteratively.

To learn knowledge from large-scale existing plans, deep supervised learning has been widely applied to \textit{dose objectives definition} and \textit{optimal fluence prediction} modules (details in Section \ref{sec:related}). 
\textit{Leaf sequencing} is relatively challenging with supervised learning because its output space is large in the serial nature,  especially for VMAT. For example, a typical HD-120 VMAT plan with 178 control points (CPs) \cite{bergman2014monte} has over 20K degrees of serial actions from leaf positions and monitor units.

Apart from being unable to leverage learning from existing data, the absence of a differentiable leaf sequencing module poses a challenge for training the overall end-to-end planning pipeline. In the spirit of recent success in decision-making tasks, we propose a new deep reinforcement learning (DRL) model, termed as \textit{Reinforced Leaf Sequencer} (RLS), for a learning-based leaf sequencing.  Our major contributions can be summarized as below: 

\begin{itemize}
    \item To our best knowledge, we are the first to successfully formulate \textit{leaf sequencing} problems in practical radiotherapy scenarios with a deep multi-agent RL model. 
    \item We propose five reward components to reasonably guide the movement of leaves and monitor units. The move pattern is controllable by tuning the weight of rewards, which enables human preferences in the loop.  
    \item  Compare to vanilla RL or optimization methods, RLS abandons iterative processing and only executes once for each CP for accelerated inference. 
    \item Experiments conducted on four datasets across two cancer sites, RLS achieved 1) better or close performances compared to a leading  optimization sequencer in a VMAT optimization platform (i.e., PORIx mentioned later); 2) effective prediction in a full-AI pipeline. 
\end{itemize}

\section{Related Work}
\label{sec:related}
\textbf{Leaf Sequencing in RTP Optimization}. RTP is mainly solved by optimization-based methods currently in both IMRT and VMAT \cite{xia1998multileaf,shepard2002direct,earl2003inverse,ripsman2022robust,carrasqueira2023automated,fallahi2022discrete,jhanwar2023portpy,cedric2011intensity}. There are different pipelines of RTP optimization, one major branch (as in Figure \ref{fig:PTP}), such as \citet{cao2006continuous,shepard2007arc,wang2008arc,bedford2009treatment}, is including the leaf sequencing to convert the optimized beam intensities into deliverable multileaf collimator (MLC) segments to form arc(s) or fields by starting from fluence maps. RapidArc$^{\text{TM}}$ from Varian is a representative commercial example in this branch which initially adopted the algorithm in \citet{otto2008volumetric} and continually upgraded in subsequent years. RapidArc has directly promoted the large-scale clinical implementation of VMAT \cite{cedric2011intensity,infusino2015clinical}. Compared to machine learning,  disadvantages of optimization algorithms include time-consuming iterative execution and the lack of leveraging knowledge from large-scale training data. 

\textbf{AI for RTP}. Recently, researchers aim to translate artificial intelligence (AI) to RTP to achieve fast and high-quality planning \cite{huynh2020artificial,luchini2022artificial} to replace one or more components of an optimization pipeline. To replace conventional point-based objectives, RapidPlan$^{\text{TM}}$ introduce dose-volume histograms (DVHs) proposal as line-objectives \cite{Fogliata2019RapidPlanStrategies,RapidPlanVarian}. Dose prediction from patient data (e.g., CT and contouring) with deep learning may serve as 3D objectives \cite{Barragan-Montero2019Three-dimensionalConfigurations,Kearney2020DoseGAN:Generation,Wang2022DeepDecomposition,gao2023flexible,feng2023diffdp,jiao2023transdose,gronberg2023deep}. Following dose prediction, deep learning methods are used to replace fluence map optimization \cite{romeijn2003novel,romeijn2004unifying} to predict 2D fluence maps for each beam angle by feeding projections from predicted 3D dose  \cite{Wang2020FluenceTherapy,Wang2021DeepBoost,ma2020deep,yuan2022accelerate}.  

\textbf{Deep RL and Multi-Agent}. Deep RL has made remarkable progress in sequential decision-making tasks, including strategy games \cite{mnih2013playing,silver2016mastering}, robotic control \cite{mnih2015human,schulman2017proximal}, autonomous driving \cite{kiran2021deep}, and landmark detection \cite{ghesu2017multi}. 
Many of these applications involve multiple agents, necessitating a systematic approach to modeling as multi-agent RL (MARL) \cite{zhang2021multi}.  
Compared to value-based RLs \cite{mnih2015human,coulom2006efficient}, policy-based methods directly explore the policy space, with better convergence
guarantees \cite{konda1999actor,yang2018finite,wang2019neural,zhang2021multi}.
Actor-critic algorithms in policy-based branch, such as PPO \cite{schulman2017proximal} and SAC \cite{Haarnoja2018SoftActor}, have achieved state-of-the-art performance in many applications. 
PPO showed surprising effectiveness in cooperative multi-agent games \cite{yu2022surprising}.

RL has been applied to RTP tasks. \citet{hrinivich2020artificial,ganeshan2021reinforcement} explored  DQN for planning, however, these studies lie in single-leaf-pair simulations. A concurrent work \cite{hrinivich2024clinical} extended \citet{hrinivich2020artificial} to enable VMAT in a 3D beam context. Differently, we target modularizing RTP pipeline to be more flexible; making leaf sequencing can leverage advances of other AI models (e.g., dose/fluence predictions) and enabling integration within optimization loops. 
\textit{To the best of our knowledge, the proposed RLS is the first MARL-based leaf sequencer in practical RTP scenarios (i.e., outputs include leaf positions at each leaf pair and monitor unit for each control point)}.

\section{Problem Formulation and Intuition} 
\label{sec:problem}
This paper focuses on leaf sequencing, the final module in RTP as in Figure \ref{fig:PTP}, to achieve deliverable outcomes—a domain scarcely explored within advanced MARL. 

\begin{figure}
    \centering
    \includegraphics[width=0.45\textwidth]{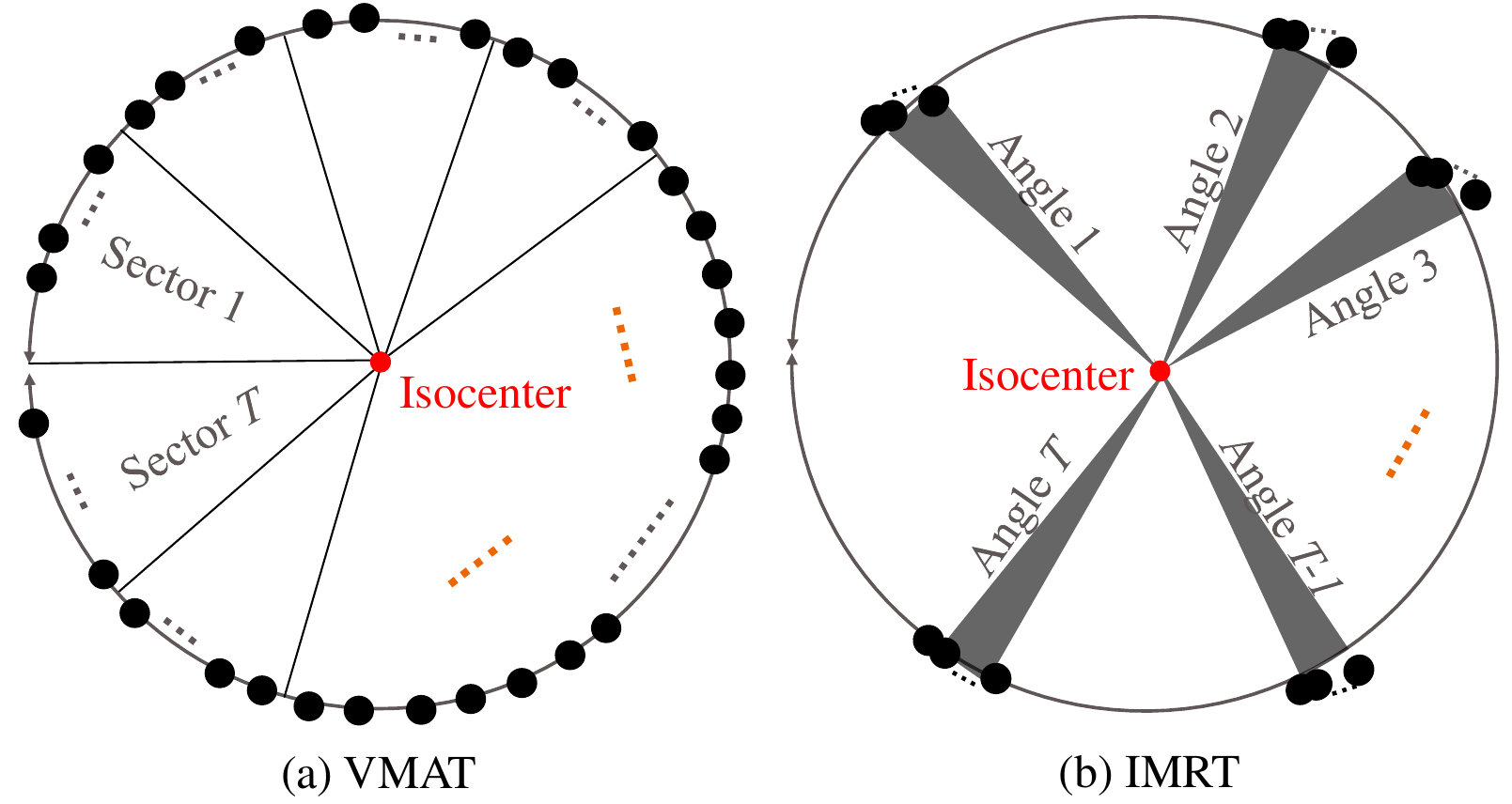}
    \caption{Control point distribution: Each black dot signifies a control point. (a) In VMAT, control points can be approximate-evenly spaced along the designated arc during therapy, and may be divided into multiple sectors during planning and optimization. (b) The IMRT plan is comprised of several fields, each encompassing multiple control points at identical gantry angles.}
    \label{fig:vmat_imrt}
\end{figure}

There are two common approaches for treatment: IMRT \cite{mundt2005intensity} and VMAT \cite{otto2008volumetric} (as in Figure \ref{fig:vmat_imrt} and more details in Appendix \ref{app:terms}). IMRT employs multiple beam-angles (i.e., fields) with modulated intensity using multileaf collimators (MLCs) to create precise dose profiles. VMAT offers improved delivery efficiency compared to IMRT, using numerous beam directions along arc trajectory to delivery dose dynamically \cite{Teh1999IntensityOncology,Teoh2011VolumetricPractice,quan2012comprehensive}. 
It is customary for IMRT to generate a sequence of control points for a specific beam angle. In VMAT scenarios, various optimization techniques, including RapidArc, address leaf sequencing by consolidating control points within a sector during optimization \cite{unkelbach2015optimization,amendola2013volumetric}. \textit{Consequently, the leaf sequencing in both VMAT and IMRT can be viewed as the process of deriving a sequence of control points, specifying leaf positions and monitor units based on target fluences, i.e., optimal fluences.}

Note that the fluence prediction component's target fluence $F$ cannot be directly translated into outcomes for the RT machine. The leaf sequencer is designed to predict outcomes that are feasible for the machine. Given $X$ leaf pairs in the MLC for all $K$ control points contributing to the fluence $F$ and denote the leaf sequencer as $f(\cdot)$, we have:
\vspace{-.05in}
\begin{equation}
    (P, M) = f(F), 
    \vspace{-.05in}
\end{equation}
where $P$ denotes the leaf positions with a size of $ K\times X \times 2$ and $M$ is a $K\times 1$ vector representing monitor unit (MU). 
\begin{figure}
    \centering
    \scriptsize
    \begin{tabular}{ccc}
    \multicolumn{3}{c}{\includegraphics[width=0.45\textwidth]{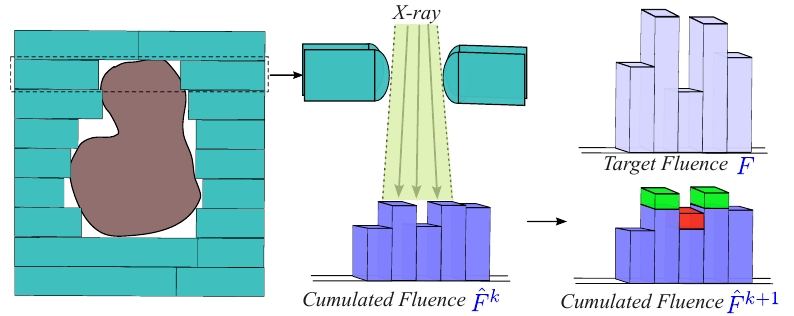}}  \\
   \hspace{0.1in} (a) multi-leaf pairs & \hspace{0.15in}(b) one-leaf pair &\hspace{0.15in} (c) one-leaf pair fluences \\
    \end{tabular}
    \vspace{-0.1in}
    \caption{(a) shows a 2D illustration of multi-leaf pairs, with the middle depicting PTV projection. (b) provides a 3D view of a leaf pair and its connection to cumulated fluences. (c) illustrates motivations of Reward 1 (green) and Reward 2 (red) by comparing cumulated and target fluences. Details are in Appendix \ref{background}.}
    \vspace{-0.1in}
    \label{fig:mlc}
\end{figure}

\begin{figure}
    \centering
    \scriptsize
    \begin{tabular}{cc}
        \multicolumn{2}{c}{\includegraphics[width=0.43\textwidth]{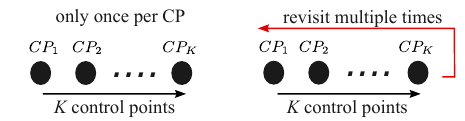}}  \\
       \hspace{0.2in} (a) RLS: finite horizon RL & \hspace{0.1in}  (b) vanilla RL or conventional methods
    \end{tabular}
    \caption{Using finite horizon RL for accelerated inference. Conventional optimization methods start with an estimate of the leaf/MU positions and iteratively refine the estimate until converge or the stopping criteria are met. In principle, we can also apply vanilla RL in an infinite horizon context, and iteratively refine the estimates. However, to achieve greater efficiency during inference, we train RLS to execute only once for each CP. }
    \vspace{-0.2in}
    \label{fig:time}
\end{figure}

\textbf{Intuition}. We summarize our intuition in Figure \ref{fig:mlc} and \ref{fig:time}. Leaf sequencing can be conceptualized as a sequence of decision-making tasks. In this sequence, the goal is to approximate target fluence with a series of control points (CPs). These decisions involve agents selecting the appropriate leaf movements and MUs for each CP. Figure \ref{fig:mlc}a shows the multi-leaf openings of one CP. Each leaf pair can filter X-rays to balance the radiation on PTV and OARs, as in Figure \ref{fig:mlc}b. By comparing cumulated and target fluences at CPs (as well as other practical regularization), a set of rewards can guide movements of leaves and MUs, as shown in Figure \ref{fig:mlc}c and Section \ref{sec:RLS}.

To reduce time complexity, we formulate leaf sequencing under a finite horizon RL adapted from \citet{pardo2018time}, where agents have to maximize expected return only over a fixed episode length, as in Figure \ref{fig:time}. To cover multiple leaf pairs and MU prediction, we leverage the multi-agent framework. Consider practical function of leaf pairs and MUs, we propose a two-level RL framework as in Figure \ref{fig:framework}.  
 
\vspace{-0.1in}
\section{Methodology: Reinforced Leaf Sequencer}
\label{sec:RLS}
\begin{figure*}
    \centering
    \includegraphics[width=0.9\textwidth]{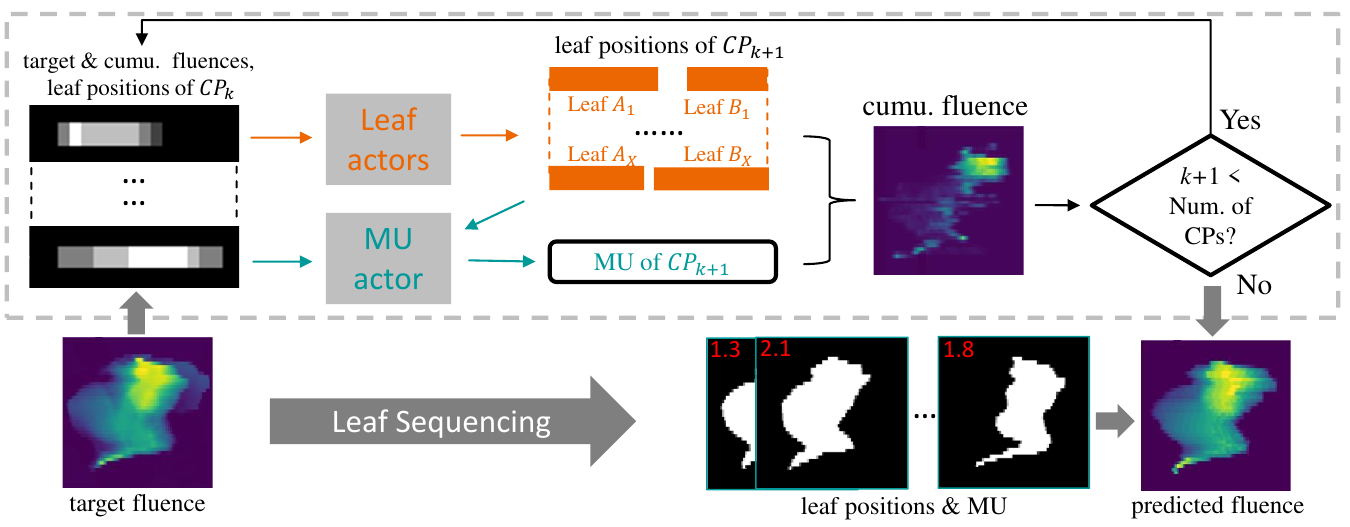}
    \caption{Illustration of the proposed RLS. The upper shows the methodology and the lower shows the input/output of RLS. The target fluence is splitted into $X$ rows, each row is related to one leaf-pair and one leaf actor. $x$-th leaf actor predicts the positions of Leaf $A_x$ and $B_x$. All rows in $k$-th control point (CP) shares the same monitor unit, which is predicted by MU actor after all leaf positions are obtained. The state of leaf actor at $CP_{k+1}$ includes target fluence, cumulated fluence of $CP_{1} \sim CP_{k}$, leaf positions of $CP_{k}$. The state of MU actor is similar but replace leaf positions of $CP_{k}$ with that of $CP_{k+1}$. }
    \label{fig:framework}
\end{figure*}

Our framework is shown in Figure \ref{fig:framework}. The backbone of RLS is in spirit of PPO \cite{schulman2017proximal}. We also conducted ablation studies of other actor-critic configurations.

\subsection{Key Components of RL}
\label{sec:definition}
    \textbf{Enviroment}. The environment takes the output of agents (leaf actors and MU actor) and compute the cumulated fluences. Target fluence generation and fluence map computation from leaf sequencing are based on methods in a planning environment PORIx (details in Appendix \ref{app:PORIx}).   

\textbf{Actions}. We define the leaf movement creating the fluence map as a sequence of discrete actions $la \sim \mathds{N}^2, -S \leq la \leq S$. $S$ is the maximum step size of each leaf. The MU action $ma$ is continuous in the range of $(R_b, R_e)$ where $R_e$ can be the maximum MU allowed by the radiotherapy machine. In $k$-th control point $CP_k$, $x$-th leaf pair has a separate leaf movement action $la_{x,k}$, while all leaf pairs of $CP_k$ share the same MU action $ma_k$. 

\textbf{State}. The state $s_k$ of the leaf actors at $CP_k$ is based on following information: target fluence $F$, cumulated fluence $\hat{F}^k$, and leaf positions $(A^k, B^k)$, the step index $k$, and leaf position index $x$. Since MU actor is to predict the optimal MU, whose state should be updated with new leaf positions after leaf actors take actions. This is the so-call \textit{two-level} actions motivated from clinical practice. For simplicity, we use $(s_k,la_k)$ to represent the state for MU actor, as in the afterwards Eq. \ref{eq:ratio}. 

\textbf{Rewards}. Constructing the reward system is pivotal for framing the leaf sequencing with RL. We define the cumulated fluence $\hat{F}^k$ as the cumulation of fluences of control points $CP_1$ to $CP_k$, i.e., $\hat{F}^{k} = \sum_{i=1}^k M^k \cdot\tilde{F}^k$, where $M^k$ and $\tilde{F}^k$ are MU and unit fluence of $CP_k$ respectively. Considerations for reward design are as follows: \textit{\textbf{1)} Encouragement to approach target fluence (Reward 1):}  Positive rewards for actions bringing cumulative fluence closer to the target, seen in the green of Figure \ref{fig:mlc}c. \textit{\textbf{2)} Punishing overdosing (Reward 2):} Penalties for actions causing cumulative fluence to exceed target intensity, represented by the red portion in Figure \ref{fig:mlc}c. \textit{\textbf{3)} Avoiding cross leaves (Reward 3):} Encouraging avoidance of left-right leaf intersection, considering its impracticality. \textit{\textbf{4)} Leaf/MU changing regularization between $CP$s (Reward 4):} Regularizing machine movement between control points, with user-adjustable parameter $\lambda_4$ in training. \textit{\textbf{5)} Aperture regularization (Reward 5)}: Regularizing the shape of the aperture formed by $(A^k, B^k)$, with area and perimeter denoted as $\text{{\fontfamily{qcr}\selectfont area}}$ and $\text{{\fontfamily{qcr}\selectfont peri}}$ respectively, which aligns with \citet{younge2012penalization}.

Rewards of $k$-th $CP$ for $x$-th leaf pair are defined as below:
\vspace{-0.1in}
\begin{equation}
\begin{split}
    R_1^k = & \sum^Y_{y=1} M^k \cdot \tilde{F}^k_{x,y} \cdot \mathds{1}(F_{x,y} - \hat{F}^{k-1}_{x,y} > M^k)  \\
    \vspace{-0.05in}
    R_2^k = & - \sum^Y_{y=1} \mathds{1}(F_{x,y} - \hat{F}^k_{x,y} < M^k) \\
    R_3^k = & \mathds{1}(B^k_x - A^k_x) - \mathds{1}(A^k_x - B^k_x)  \\ 
    R_4^k = &\Sigma_{I \in \{A,B,M\}} (1 -  \sigma(|I^k_x-I^{k-1}_x|)) \\ %
    R_5^k = & \text{{\fontfamily{qcr}\selectfont area}}(A^k, B^k) / \text{{\fontfamily{qcr}\selectfont peri}}(A^k, B^k)
\end{split}
\label{eq:rewards}
\end{equation}
where $Y$ is the second dimension size of fluence $F$. The total reward is defined as $R^k = \lambda_1\cdot R_1^k + \lambda_2\cdot R_2^k + \lambda_3\cdot R_3^k + \lambda_4\cdot R_4^k + \lambda_5\cdot R_5^k$. Details with pseudo-code are shown in Appendix \ref{rw_code} and math symbol summary is in Appendix \ref{background}.

\subsection{Two-level Multi-agent PPO}

The proposed RLS adapts a multi-agent PPO framework. To accommodate RT characteristics, it incorporates two levels of actions: leaf movements and MUs prediction. RLS trains three networks: leaf policy net $\pi_{\theta_l}$ shared by all $X$ leaf agents parameterized by $\theta_l$, MU policy $\pi_{\theta_m}$ for MU agent parameterized by $\theta_m$, and critic net $V_{\phi}$ parameterized by $\phi$. 

The policy nets are trained to maximize the objective: 
\begin{equation}
L(\theta) = L^{\text{CLIP}}_{\text{sur}}(\theta) + c  \hat{\mathbb{E}}_{x,k} \left[ H(\pi_{\theta_l}) \right] + c\hat{\mathbb{E}}_{k} \left[ H(\pi_{\theta_m}) \right],
\label{eq:policy_loss}
\end{equation}
where $H(\pi_\theta)$ is the entropy of the policy distribution, and $c$ is the entropy coefficient hyperparameter. The clipped surrogate objective $L^{\text{CLIP}}_{\text{sur}}(\theta)$ is given by:
\begin{equation}
\scalebox{0.8}{$
L^{\text{CLIP}}_{\text{sur}}(\theta) = \hat{\mathbb{E}}_{x,k} \left[ \min \left( r_{x,k}(\theta) \hat{A}_{x,k}, \text{clip}\left(r_{x,k}(\theta), 1 \pm \epsilon \right) \hat{A}_{x,k} \right)  \right]$},
\label{eq:ppo_surrogate_loss}
\end{equation}
where the advantages $\hat{A}_{x,k}$ are computed using GAE \cite{schulman2015high} with rewards definition in Section \ref{sec:definition}. The probability ratio  $r_{x,k}(\theta)$ for $x$-th leaf pair of $CP_k$ is 
\begin{equation}
\label{eq:ratio}
r_{x,k}(\theta) = \frac{\pi_{\theta_l}(la_{x,k}|s_{x,k}) \cdot \pi_{\theta_m}(ma_{k}|s_{k},la_{x,k})}{\pi_{\theta_{l,\text{old}}}(la_{x,k}|s_{x,k}) \cdot \pi_{\theta_{m,\text{old}}}(ma_{k}|s_{k},la_{x,k})}.
\end{equation}

The critic net is trained to minimize the loss function (with clipping following CleanRL \cite{huang2022cleanrl}):
\begin{equation}
\scalebox{0.9}{$
L(\phi) = \hat{\mathbb{E}}_{x,k} \left[ \frac{1}{2} \left(V_{\phi}(s_{x,k}) - \left(r_{x,k} + \gamma V_{\phi}(s_{x,k+1})\right)\right)^2 \right]$},
\label{eq:value}
\end{equation}
The RLS training algorithm is shown in Algo. \ref{alg:example}.

\begin{algorithm}[tb]
\small
\caption{RLS training}\label{alg:example}
\begin{algorithmic}
\STATE {\bfseries Input:} $X$ leaf agents with shared policy $\pi_{\theta_l}$ and an MU agent with policy $\pi_{\theta_m}$ 
\STATE {\bfseries Output:} trained leaf policy $\pi_{\theta_l}$, MU policy $\pi_{\theta_m}$, critic $V_{\phi}$
\end{algorithmic}
\begin{algorithmic}[1]
   \STATE Initialize parameters $\{\theta_l, \theta_m\}$ for policies $\{\pi_{\theta_l}, \pi_{\theta_m}\}$ and parameters ${\phi}$ for critic $V_{\phi}$
   \FOR{$n$ = 1 to \textit{RL iterations}}
   \STATE Initialize data buffer $D$
    \FOR{$i$ = 1 to \textit{batch size}}
    \STATE Run policy $\pi_{\theta_{l, old}}$ in the environment for $X$ leaf agents 
    \STATE Run policy $\pi_{\theta_{m, old}}$ in the environment for MU agent
    \STATE Compute the advantages using GAE
    \STATE Adding timestamp data to buffer $D$    
    \ENDFOR
   \FOR{$e$ = 1 to \textit{update epochs}}
    \FOR{$i$ = 1 to \textit{batch size}}
    \STATE Compute the policy loss based on Eq. \ref{eq:policy_loss}
    \STATE Compute the value loss based on Eq. \ref{eq:value}
    \STATE AdamW update $\{\theta_l, \theta_m\}$ on $\{\pi_{\theta_l}, \pi_{\theta_m}\}$ like PPO
    \STATE AdamW update ${\phi}$ on $\pi_{\theta_m}$ like PPO
    \ENDFOR 
   \ENDFOR
   \STATE $\theta_{l, old} \leftarrow \theta_{l}$
   \STATE $\theta_{m, old} \leftarrow \theta_{m}$
 \ENDFOR
\end{algorithmic}
\end{algorithm}

\subsection{Dealing with Heterogeneous Fluence Patterns}
\begin{figure}
    \centering
    \includegraphics[width=0.48\textwidth]{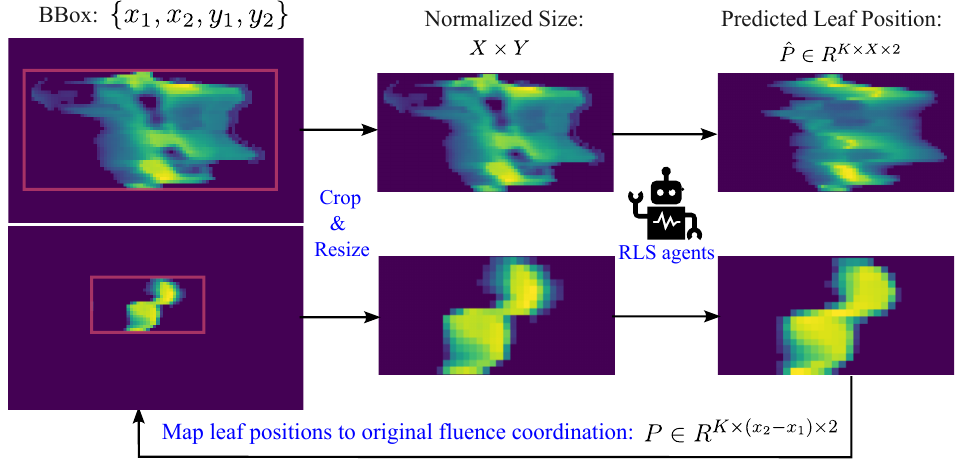}
    \vspace{-0.1in}
    \caption{Illustration of dealing with heterogeneous fluences with proposed \textit{cropping} strategy. The red box is ROI covers fluence locations with positive intensity. The RLS agents module only deal with normalized fluence maps.}
    \label{fig:heter}
    \vspace{-0.1in}
\end{figure}
As shown in Figure \ref{fig:heter}, even if  target fluences are from the same site (e.g., head-and-neck), their pattern can be significantly different due to patient distinction (e.g., different size and shape of PTV). To reduce such heterogeneity, we propose a \textit{cropping} strategy to normalize the target fluence.  

As in Figure \ref{fig:heter}, we first detect positive values in target fluence with coordination $\{\hat{x_1},\hat{x_2},\hat{y_1},\hat{y_2}\}$,
and crop the target fluence with bounding box $\{x_1 =\hat{x_1},x_2 = \hat{x_2}, y_1 = \hat{y_1} / 2, y_2 = (\hat{y_2} + Y)/2\}$.
In the training, $y_1$ and $y_2$ are sampled from $(0, \hat{y_1})$ and $(\hat{y_2}, Y)$ respectively for data augmentation. $\hat{P}$ is predicted leaf positions in the output space of RLS, which is transformed to original fluence coordination as $P$  obtained by linear interpolation and resize $P=\text{Resize}(\hat{P} \cdot \frac{y_2 - y_1}{Y} + y_1,(K, x_2 - x_1, 2))$. 

\subsection{Dealing with Various Length of Sector}

Due to various of machine or optimization settings, the number of control points in one sector may be different. To make a single model that can be applied to various lengths, we provide a novel post-processing strategy: In order to merge two (for example) control points presented as coordinates $p_{k} = p_{k-1} + d_k$, and $p_{k+1}=p_k + d_{k+1}$ (where $p_k$ and $d_k$ are the position and predicted position change associated to $CP_k$); the merged position at $CP_k$ is defined as $\bar{p}_{k} = \frac{p_{k} + p_{k+1}}{2} = p_{k-1} + d_{k} + \frac{d_{k+1}}{2}$. Thus, RLS can deal with various number of CPs in a sector with post-processing and without fundamentally change the RL module.

\subsection{Innovation Highlight}

In this study, we apply PPO and multi-agent learning principles to develop an innovative RL model that incorporates two-level agents (leaf agents and MU agent) in a finite horizon context, aiming to reduce time complexity. This model is tailored to address real-world challenges in RL. Our proposed reward mechanism empowers controllable movement patterns of the leaf sequence, a crucial element in RTP. Our study demonstrates an inaugural application of deep multi-agent RL methods of leaf sequencing.  This approach has the potential to partially replace optimization methods in RTP and facilitate end-to-end learning with other AI modules. 

\section{Experiment}

\subsection{Dataset,  Pre-processing, and Plan Evaluation}
Two sites of RT cancer treatment from four datasets are included: head-and-neck (HN) and prostate (Pros). The HN site includes three datasets: 1) the HNd set contains 493 patients after quality assurance used for training and test with a train/validation/test split, 2) two external test sites from TCIA: HNe1 \cite{TCIA1} with 31 patients and HNe2 \cite{TCIA2} with 140 patients after filtering. The Pros site is from a public dataset with access permission requirements, including 555 patients after filtering. Each patient has up to 525 target fluences from PORIx environment. Details are in Appendix \ref{app:data}. 

We have three contexts to evaluate the proposed model. 1) Photon Optimizer Research Inference (PORIx) of a leading commercial radiotherapy company Varian; 2) end-to-end full-AI research pipeline for VMAT; and 3) simulation with IMRT. 
The PORIx is a leading optimization environment 
 which provides callbacks of target fluence and leaf positions/MUs. It also allows the user to overwrite the output of the leaf sequencer, which can compare our method with the leaf sequencer in PORIx from multiple perspectives (see Appendix \ref{app:PORIx}). The leaf sequencing optimization in PORIx, terms as PORIx for short in some contexts, serves as a baseline to compare with our RLS. 

 \subsection{Evaluation Metrics} 
 We include three levels of evaluation for leaf sequencing: 
 \textbf{(1)} Mean of Normalized Square Error (MNSE) to evaluate the fluence reconstruction performance, which is defined as $\frac{1}{n}\sum_{i=1}^n\frac{||F-\hat{F}||_2}{||F||_2}$. 
 \textbf{(2)} Iterations Reducing $p$\%  Error (IRE$p$\%) to evaluate the optimization convergence speed of PORIx when its leaf sequencer is using its default model vs. our RLS. Specifically, two types of IRE are defined: The \textit{absolute}
IRE$p$\% (AIRE$p$\%) is defined as the smallest of iteration whose planning cost is smaller than $E_{1} \cdot (1 - p\%)$, where $E_{1}$ is the cost after the first iteration in PORIx. The \textit{relative}
IRE$p$\% (RIRE$p$\%) is defined as $ (E_{1} - \min(E)) \cdot (1 - p\%) + \min(E)$. The list $E$ is costs from the planning costs of iterations in PORIx. 
\textbf{(3)} To evaluate the reconstruction performance from 3D dose perspective (3D doses are computed from target vs. predicted fluences), we use the Dose score and DVH score from the OpenKBP challenge \cite{Babier2020OpenKBP:Challenge}. 

\subsection{Experimental Settings} 
Our implementation is motivated by CleanRL \cite{huang2022cleanrl}.  Our major backbone is based on PPO, which has shown success in many applications including GPT-4 \cite{openai2023gpt4}. The actions of leaves/MUs follow the physical regularization in the planning system, and MU actions of a sector/field are refined with ridge regression. We follow hyper-parameters related to PPO settings in CleanRL unless mentioned. We apply the AdamW optimizer with an initial learning rate of 1e-4, weight decay 1e-4, and a Cosine Annealing scheduler.  The reward ratios $\{\lambda_i\}$ are set to $\{1, 2, 2, 1, 1\}$ except in ablation studies. The action hyperparameters $S$, $R_b$, $R_e$ are set as 4, 0.5, 2.5, respectively. More details can be found in Appendix \ref{app:param}. 

\subsection{Main Experimental Results}
\begin{figure}
    \centering
    \begin{tabular}{ccc}
       \hspace{0.2in} (a) target  &\hspace{0.4in}  (b) ours & \hspace{0.2in} (c) PORIx \\
       \multicolumn{3}{c}{\includegraphics[width=0.48\textwidth]{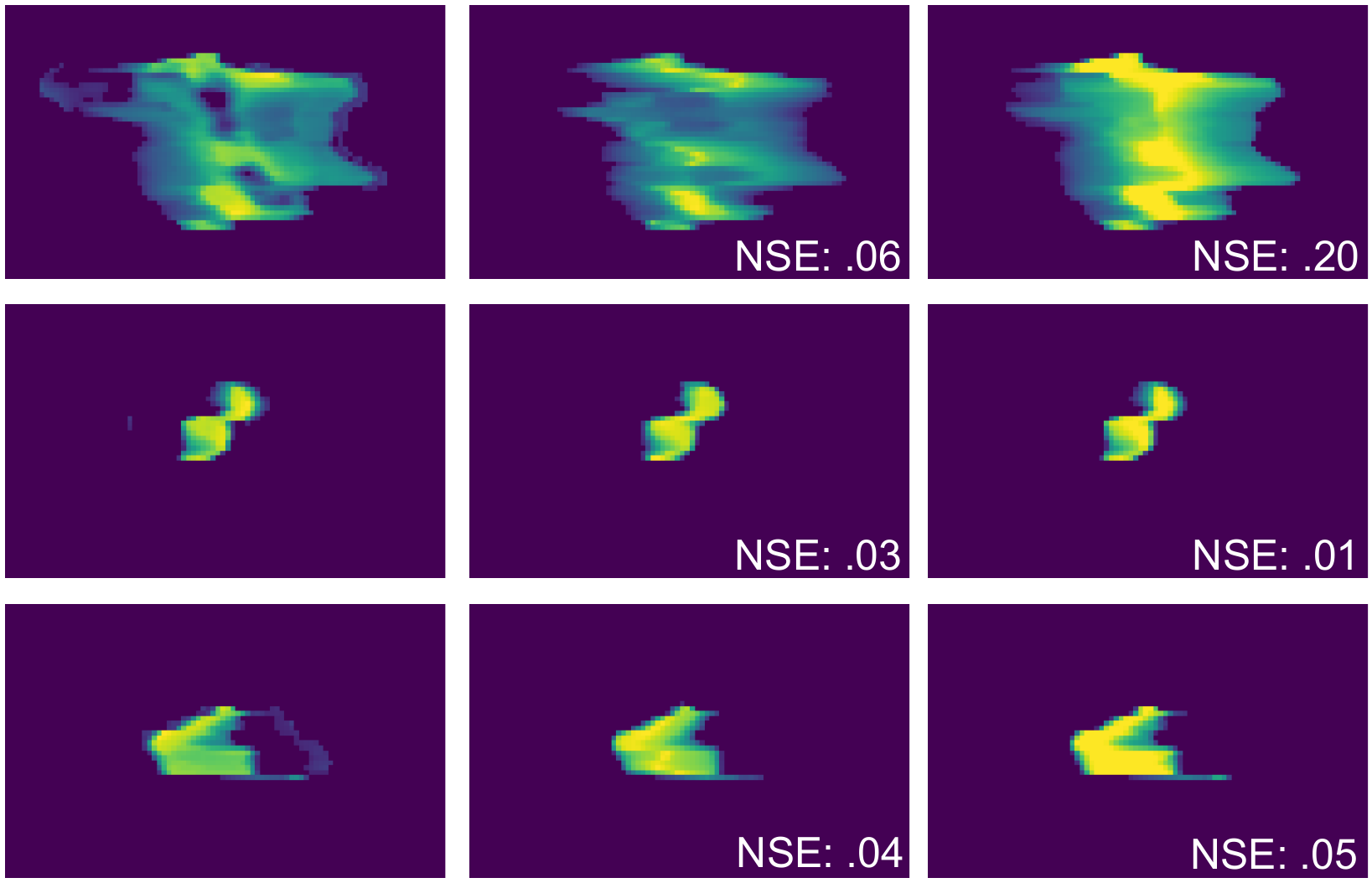}} 
    \end{tabular}
    \caption{Predicted fluence vs. target fluence. The reconstruction error is shown right bottom of the predicted fluences. First row: hard case from HN; second row: easy case from HN; third row: prostate case. This figure is matched with cases in Figure \ref{fig:ptv_cases}.}
    \label{fig:example_fluence}
\end{figure}

\begin{figure}
    \centering
    \includegraphics[width=0.45\textwidth]{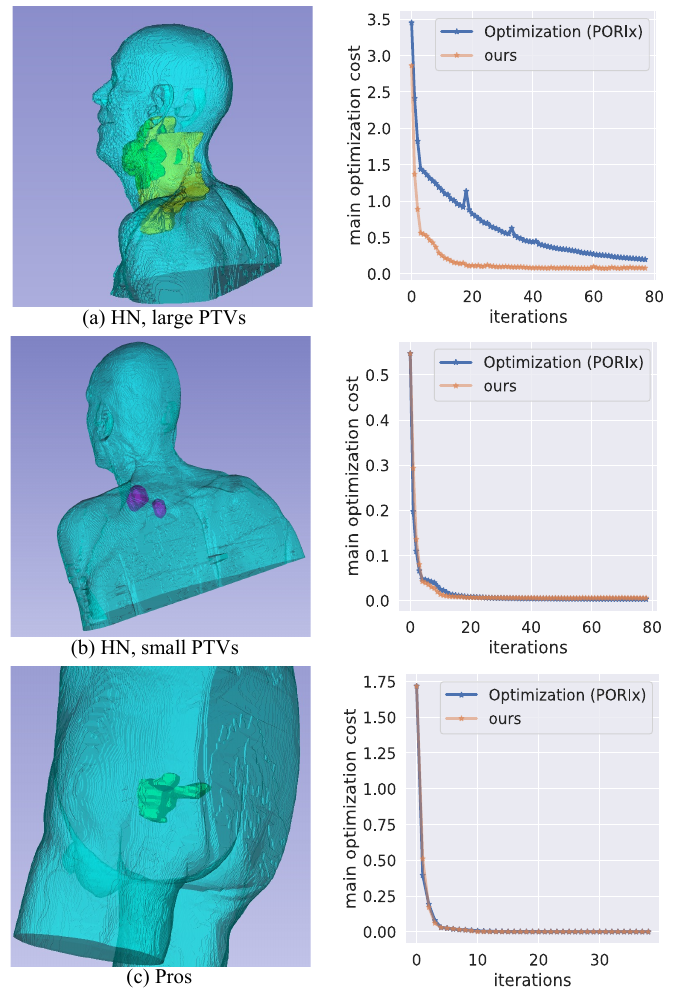}
    \caption{Typical cases in different scenarios. The left panel shows the PTV contours within body mask; right panel shows the main optimization cost with the number of iterations. Upper: large PTVs (PTV 54 and PTV 60) in HN, middle: small PTVs in HN (PTV 66), lower: PTV in prostate. The RLS brings clear improvements for hard cases (e.g., those with large PTVs).}
    \vspace{-0.1in}
    \label{fig:ptv_cases}
\end{figure}

\textbf{Planning with PORIx}. Table \ref{tab:rec_err} and \ref{tab:iter_num} show the comparison between our RLS and leaf sequencing optimization in PORIx. We achieve the lower MNSE across all compared contexts, indicating our RLS can reconstruct target fluence better than PORIx with executable leaf positions and MUs. The reduced AIRE and RIRE suggest that an optimization planner, such as PORIx, incorporating our RLS, has the potential for quicker convergence.
\begin{table}[t]
    \centering
    \small
    \begin{tabular}{ccccccc}
    \toprule
         & HNd & HNe1 & HNe2 & Pros  & Pros(e) \\
    \midrule
     PORIx   & .219  &   .257  &  .241     & .079  & .079 \\
     ours   & \textbf{.149}  &     \textbf{.165} & \textbf{.146}     &  \textbf{.042} & \textbf{.043}  \\
     \bottomrule
    \end{tabular}
    \caption{The MNSE ($\downarrow$) from target fluence and predicted fluence.}
    \label{tab:rec_err}
\end{table}
\begin{table}[]
    \centering
    \small
    \begin{tabular}{cccccc}
    \toprule
         & HNd & HNe1 & HNe2 & Pros  & Pros(e)  \\
    \midrule
    \multicolumn{6}{c}{\textcolor{blue}{RIRE97\% metric ($\downarrow$)}} \\
     \hdashline
     PORIx   &  18.8 &   26.3  &   20.6  & 4.13 &  4.13 \\
     ours  & \textbf{12.6}  &  \textbf{10.7}  & \textbf{14.4}  &  \textbf{3.85} &  \textbf{3.75} \\
     \hdashline
     \multicolumn{6}{c}{\textcolor{blue}{AIRE97\% metric ($\downarrow$)}} \\
     \hdashline
     PORIx    &  27.2 &    46.8 &  41.0   & 4.21 & 4.21 \\
     ours   &  \textbf{17.6}   &   \textbf{27.9} &  \textbf{31.6}   & \textbf{3.93}  &  \textbf{3.78} \\
     \hdashline
     \multicolumn{6}{c}{\textcolor{blue}{AIRE99\% metric ($\downarrow$)}} \\
     \hdashline
     PORIx    &  56.7 &    71.0  &   62.2 & 8.65 & 8.65  \\
     ours   &  \textbf{50.1}   &   \textbf{66.3} &  \textbf{53.1}  & \textbf{7.25}  &  \textbf{7.05} \\
     \bottomrule
    \end{tabular}
    \caption{RIRE97\%, AIRE97\% and AIRE99\% of PORIx vs. ours.}
    \label{tab:iter_num}
\end{table}
Figure \ref{fig:example_fluence} and \ref{fig:ptv_cases} depict typical outcomes of our RLS and the PORIx optimizer. The RLS excels in challenging cases, especially in large and heterogeneous PTV cases of the head-and-neck (first row in both figures). In easier cases of head-and-neck and prostate (second and third rows), RLS performs close or slightly outperforms PORIx optimizer. Notably, easy cases exhibit better reconstruction and faster convergence, which may not pose a significant challenge in RTP. 
\begin{table}[]
    \centering
    \scriptsize
    \begin{tabular}{ccccc}
    \toprule
         &  \multicolumn{2}{c}{Dose score ($\downarrow$)}  & \multicolumn{2}{c}{DVH score ($\downarrow$)} \\
         & MAE & MSE & MAE & MSE \\
    \hline
    OpenKBP S1 & 2.4 (1st) & 15.5 (1st) & 1.5 (1st) & 5.9 (1st)\\
    OpenKBP S2 & 2.6 (2nd) & 16.6 (2nd) & 1.7 (12nd) & 6.8 (10th)\\
    OpenKBP S3 & 2.7 (4th) & 18.1 (5th) & 1.5 (2nd) & 6.0 (2nd)\\
     \citet{teng2024beam} & 2.1 & - & 0.98 &  - \\
     \hline
     \hline
     
     ours (VMAT, 8 cps)   &  0.19         & 0.18     & 0.88 & 1.2 \\
     ours (IMRT, 32 degree)   &  0.23         & 0.28     & 0.97 & 1.6 \\
     \bottomrule
    \end{tabular}
    \caption{Dose and DVH scores when comparing a 3D dose from prediction and a 3D dose from target. The upper panel is best performances in the OpenKBP leaderboard \cite{Babier2020OpenKBP:Challenge} and the latest breakthrough; the lower panel is for evaluating leaf sequencing when target fluence from fluence prediction module. } 
    \label{tab:metric}
\end{table}

\textbf{Full-AI End-to-end VMAT Planning (Prostate)}.  
As in Figure \ref{fig:PTP}, a typical RTP pipeline can consist of three major components. Here, we evaluate our RLS in the context of a full-AI end-to-end framework without optimization iteration. The objective definition is replaced by dose prediction adapted from \citet{gao2023flexible}, and the fluence prediction is motivated by \citet{Wang2020FluenceTherapy} (which was originally for IMRT field dose, here has been adapted for VMAT). More details of dose/fluence predictions are in Appendix \ref{app:othermodule}. 

As in Figure \ref{fig:aie2e} and Table \ref{tab:metric}, the computed 3D doses from target fluence (not executable for machine) and predicted fluence from RLS (executable from machine) are close, both visually and in terms of Dose/DVH scores. Compared to dose prediction (a module needed in E2E pipeline) from OpenKBP challenge \cite{Babier2020OpenKBP:Challenge}, RLS achieves better Dose score and DVH score compared to state-of-the-art models in OpenKBP \footnote{It is a compromise to use records of OpenKBP to compare with our model since our tasks are not the same as in OpenKBP.  See detailed considerations in Appendix \ref{app:othermodule}.}. 

\textbf{Simulation for IMRT}.
Although RLS was designed for VMAT contexts, it can be applied to leaf sequencing of IMRT with a simulation testbed. We group 16 control points ($\sim$32 degree between two adjacent beam angles) as a single field from fluence prediction module of the end-to-end pipeline in a IMRT-simulation context. As in Table \ref{tab:metric} (last row), our RLS also achieves promising performance.

\begin{figure}
    \centering
    \begin{tabular}{ccc}
    \hspace{-0.15in}(a) target & \hspace{-0.28in} (b) prediction & \hspace{-0.25in} (c) difference \\
    \hspace{-0.1in}\includegraphics[width=2.5cm]{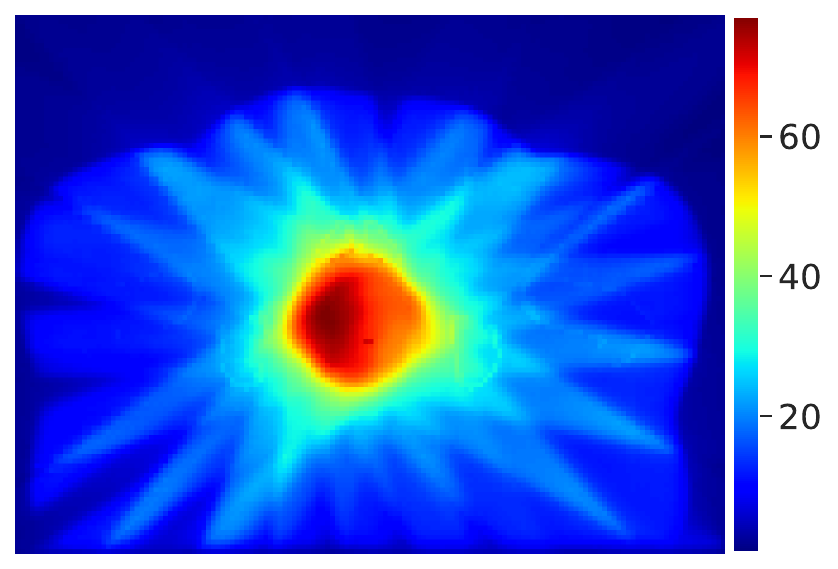}& \hspace{-0.2in}
\includegraphics[width=2.5cm]{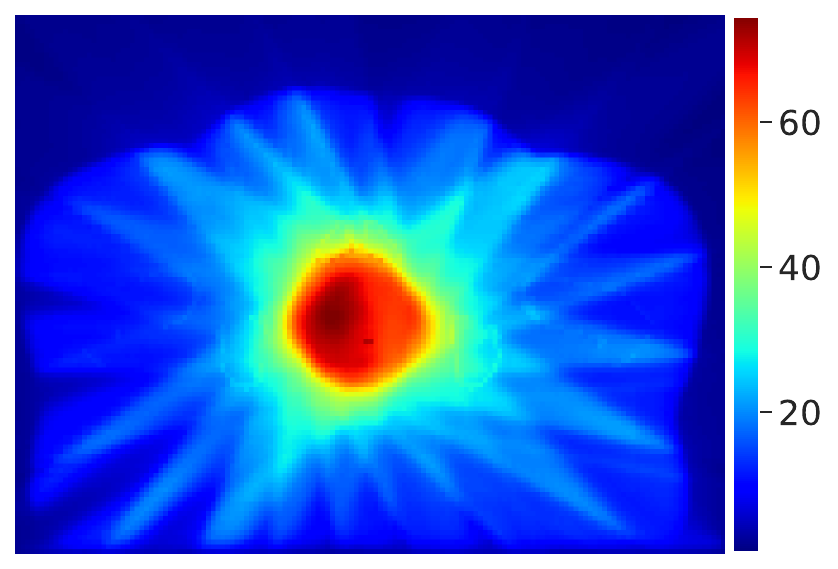} & \hspace{-0.2in} \includegraphics[width=2.5cm]{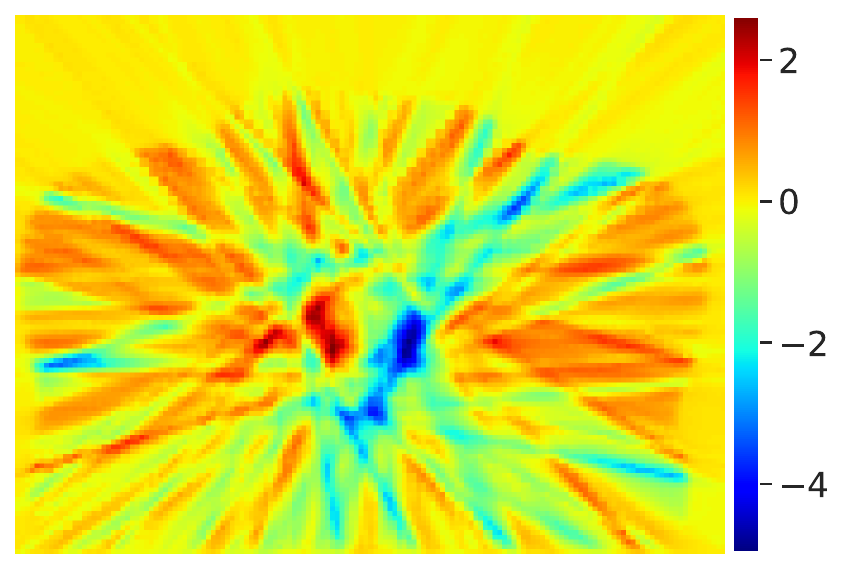} \\
\multicolumn{3}{c}{\includegraphics[width=6cm]{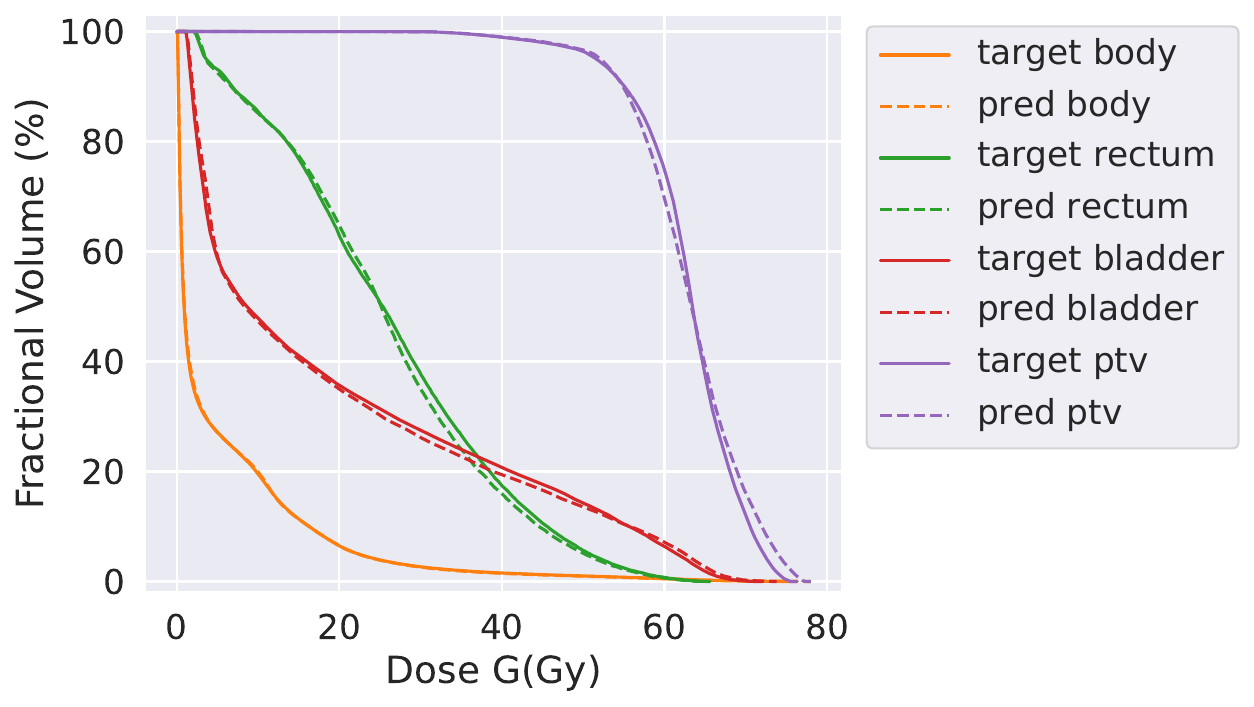}}
    \end{tabular}
    \caption{RT plan from an AI end-to-end preliminary solution. (a) \textit{target} is the 3D dose computed from fluences (not machine executable) of \textit{Fluence Prediction} module, while (b) \textit{prediction} is computed from executable leaf sequencing. The lower part shows DVHs from \textit{target} (solid lines) and \textit{prediction} (dash lines).}
    \label{fig:aie2e}
\end{figure}

\subsection{Ablation Studies}
\label{sec:ablation}
\begin{figure}
    \centering
    \includegraphics[width=0.45\textwidth]{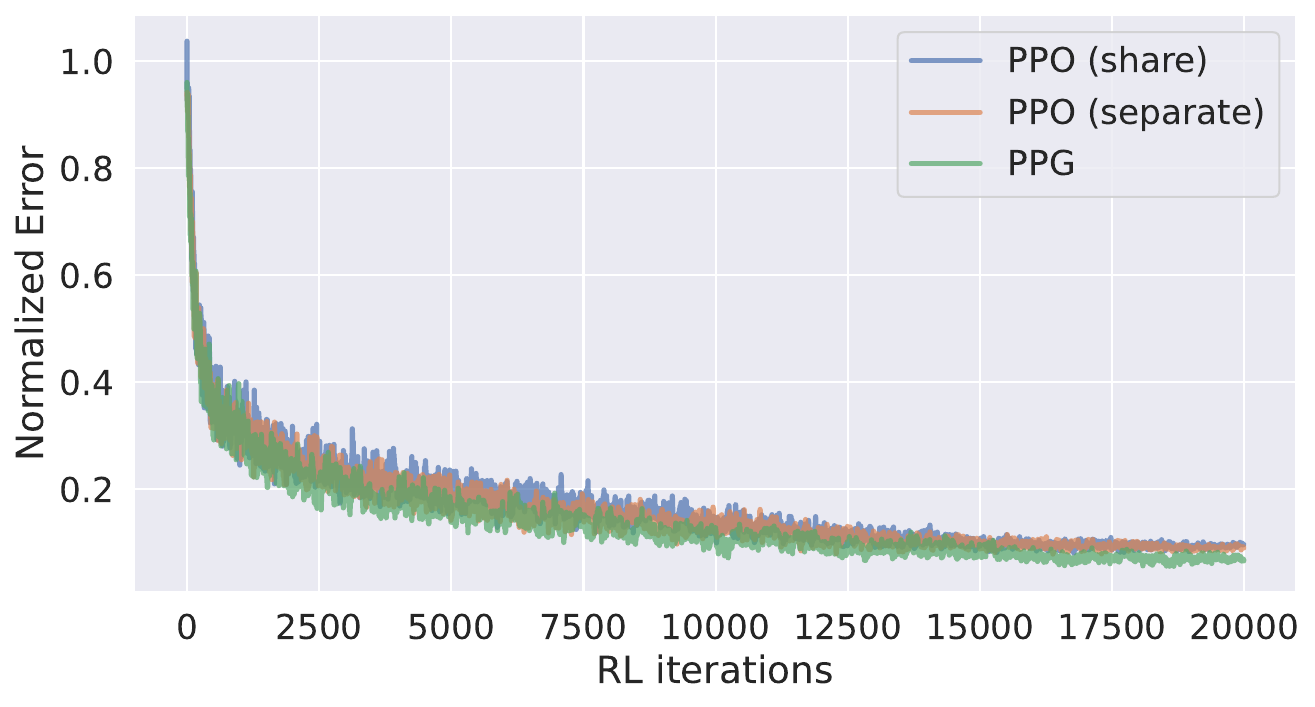}
    \caption{The ablation of different actor-critic structures with normalized reconstruction errors. \textit{share} and \textit{separate} represent the policy-net and critic-net are shared and separated, respectively.}
    \label{fig:ppg_ppo}
\end{figure}
\textbf{Backbone}. Our major experiments follow the settings of PPO in CleanRL \cite{huang2022cleanrl}, i.e., the networks of actor and critic are separated. We include the ablation studies of sharing backbone between actor and critic, and PPG \cite{cobbe2021phasic} which separates actor and critic training into distinct phases. As shown in Figure \ref{fig:ppg_ppo}, in our leaf sequencing context, the shared-network setting is a little worse than the separate-network under the PPO setting. Overall, the PPG framework has minor superiority over PPO backbones. Considering differences are not significant in these three backbones and for simplicity purposes, we apply the most widely used PPO (separated) for the majority of our experiments. We leave careful comparisons of different backbones in future work. 
 \begin{figure}
    \centering
    \begin{tabular}{cc}
\hspace{-0.1in}\includegraphics[width=4.2cm]{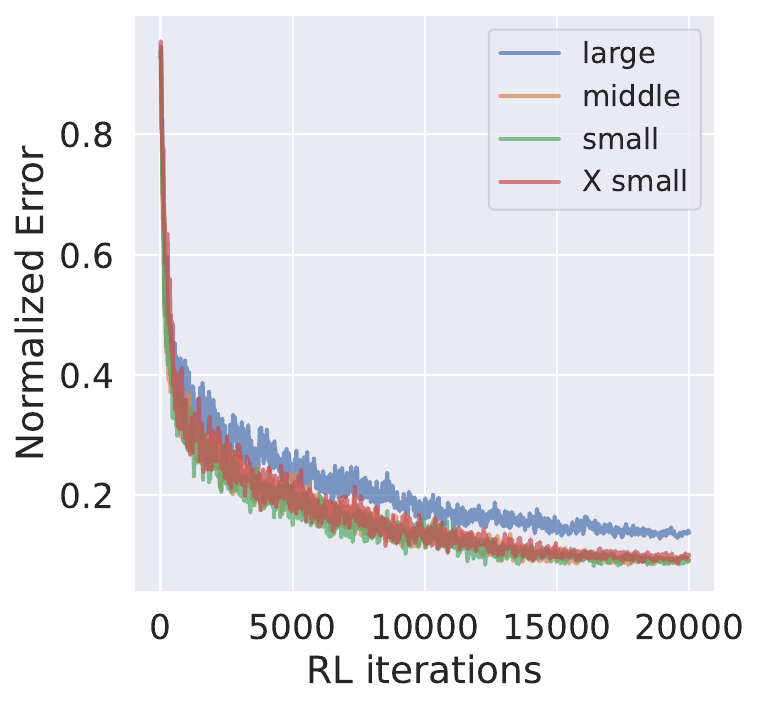}& \hspace{-0.2in}
\includegraphics[width=4.2cm]{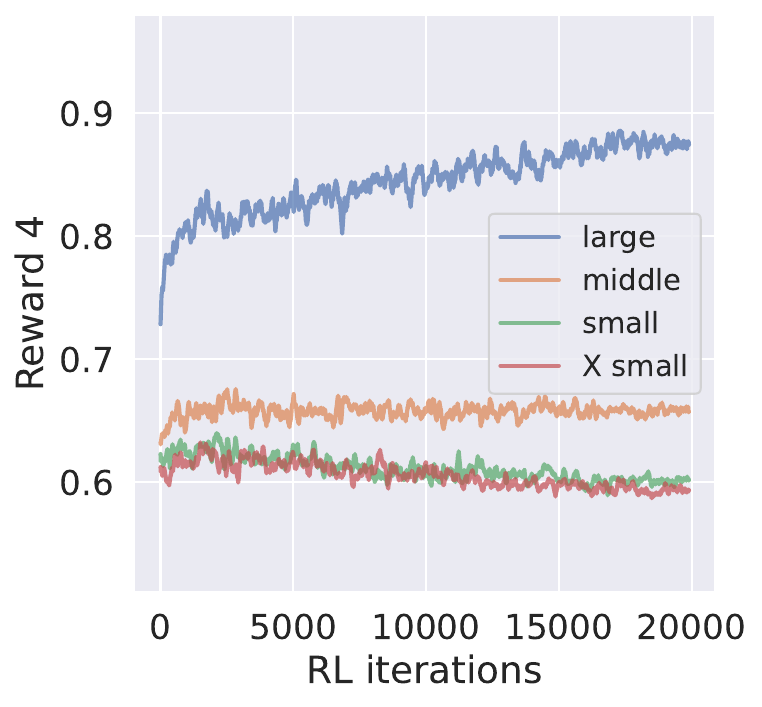}\\
\vspace{-0.05in}
(a) Normalized Errors & (b) Reward 4  \\
\vspace{-0.15in}
\end{tabular}
    \caption{Different weights on leaf speed regularization (i.e., Reward 4 in Eq. \ref{eq:rewards}). Four scales of $\lambda_4$ are compared; Large: 5, middle: 1, small: 0.1, X small: 0.01.}
    \label{fig:reward4}
\end{figure}

\textbf{The Ratio of Reward Components}. 
To better understand the sensitivity of different reward components, we conduct the ablation studies with five different ratios of each reward, as shown in Table \ref{tab:app_rw_abl}. 

\begin{table}[h]
    \centering
    \begin{tabular}{cccccc}
    \toprule
        Weight & 0 & 0.01 & 0.1 & 1 & 10 \\
        \midrule
        $\lambda_1$ & 0.300 & 0.289 & 0.229 & 0.149 & 0.181 \\
        $\lambda_2$ & 0.574 & 0.528 & 0.271 & 0.151 & 0.226 \\
        $\lambda_3$ & 0.155 & 0.152 & 0.150 & 0.149 & 0.151\\
        $\lambda_4$ & 0.148 & 0.147 & 0.147 & 0.149 & 0.167 \\
        $\lambda_5$ & 0.150 & 0.148 & 0.149 & 0.149 & 0.165 \\
    \bottomrule
    \end{tabular}
    \caption{The reconstruction errors (MNSE) of five weights in different scale for each reward ratio $\lambda_i$ (other $\lambda_{j\neq i}$) are set to default ratio. The default ratios are $\{\lambda_1:1, \lambda_2:2, \lambda_3:2, \lambda_4:1, \lambda_5:1 \}$, which achieve 0.149 of MNSE. }
    \label{tab:app_rw_abl}
\end{table}

 Reward 1 and 2 are the major rewards driving the feasibility of RLS, as depicted in Figure \ref{fig:app_mlc}c, which are more sensitive to weights. Reward 3-5 are auxiliary with clinical or physical considerations, which are less sensitive to the weight. The default weights achieved a reasonably good performance, exposing such ablation studies guides users to choose weights in different scenarios.

Figure \ref{fig:reward4} shows a detailed ablation study of the leaf change in adjacent control points (i.e., $R^k_4$ in Eq. \ref{eq:rewards}).  High leaf speed may increase the instability
and violate the max speed of the RT machine, while low speed may reduce the flexibility to reconstruct the target fluence. We demonstrate that by tuning the weight of the reward (i.e., $\lambda_4$) in training, our model allows user preference in the balance between leaf speed and reconstruction performance. 

\textbf{Cropping Strategy}. The utilization of the \textit{cropping} strategy facilitates the normalization of heterogeneous fluence patterns. A comprehensive comparison, as in Table \ref{tab:crop}, highlights the impact of \textit{cropping} vs. \textit{no cropping} and the optimization solution. Significantly, the \textit{cropping} strategy has enhanced the performance of leaf sequencing, particularly evident in external testing across diverse sites, as indicated by the columns for Pros(e) in Table \ref{tab:crop}. This scenario involves training on head-and-neck cancer site and subsequently external testing on prostate cancer site. More experimental results can be found in Appendix \ref{app:crop}. 
\begin{table}[]
    \centering
    \small
    \begin{tabular}{ccccc}
    \toprule
         & \multicolumn{2}{c}{HNe1} &  \multicolumn{2}{c}{Pros(e)} \\
    \midrule
         & w/o Crop & \hspace{-0.1in}w Crop & w/o Crop & \hspace{-0.1in}w Crop \\
    \hdashline
      RIRE97\% ($\downarrow$)   & 13.2 & \textbf{10.7}\tiny$\textcolor{blue}{-19\%}$ & 6.13 & \textbf{3.85}\tiny$\textcolor{blue}{-38\%}$\\
      AIRE97\% ($\downarrow$)   &  \textbf{27.3} & 27.9\tiny$\textcolor{red}{+2\%}$& 6.25 & \textbf{3.93}\tiny$\textcolor{blue}{-37\%}$\\
      AIRE99\% ($\downarrow$)   &   69.4 & \textbf{66.3}\tiny$\textcolor{blue}{-4\%}$& 9.44  & \textbf{7.25}\tiny$\textcolor{blue}{-23\%}$\\
      MNSE  ($\downarrow$)   &  .181 &\textbf{ .165}\tiny$\textcolor{blue}{-9\%}$&  .060 & \textbf{.043}\tiny$\textcolor{blue}{-28\%}$\\
      \bottomrule
    \end{tabular}
    \caption{With vs. without cropping comparison. RLS is trained on HNd, and externally tested on HNe1 and prostate sites.}
    \vspace{-0.1in}
    \label{tab:crop}
\end{table}

\section{Discussion}

\textbf{Conclusion}. This paper introduces the \textit{Reinforced Leaf Sequencer} (RLS), a pioneer multi-agent reinforcement learning approach in practical leaf sequencing for radiotherapy. RLS can potentially supplant commonly used optimization-based methods, producing executable plans. Our technical contribution involves extending multi-agent and Proximal Policy Optimization (PPO) into a two-level practical leaf sequencing framework, incorporating novel rewards and actions. We assess the performance of the proposed RLS using four datasets (three from head-and-neck site and one from prostate site) in three distinct contexts: 1) a practical radiotherapy environment, 2) a full-AI end-to-end research pipeline, and 3) IMRT simulation. Notably, our AI model, even without an iteration loop, exhibits improvements in terms of reconstruction error and early iteration coverage rate when compared to optimization methods.

\myparagraph{Limitation and Future Work}. As an initial endeavor to address practical leaf sequencing with MARL, our work has several limitations. \textbf{\textit{First}}, unlike well-established research domains, our comparative analyses are constrained by the scarcity of directly relevant literature. Moreover, the absence of open-source models, often attributable to commercial or confidentiality constraints, further hinders direct comparisons. Fortunately, the PORIx provides a contemporary baseline for evaluation and facilitates meaningful comparisons. \textbf{\textit{Second}}, the training of RLS is focused on leaf sequencing module so that main-loop costs in PORIx may not be always well-optimized.  This limitation is similarly present in the end-to-end AI planning pipeline. One future work is to train different modules end-to-end for global optimal plan. \textbf{\textit{Third}}, we currently concentrate solely on the initial $\leq$ 100 iterations of PORIx as a preliminary plan. In clinical practice, the number of iterations can theoretically be infinite, and planners can use multi-resolution planning (as in Eclipse) to fine-tune solutions. We simplified this process for large-scale automation. One potential future work could involve initiating the planning process with deep learning plans, which offer rapid and automated solutions, followed by a limited number of human/optimization intervention steps. More contexts can be found in Appendix \ref{app:PORIx}. 
\textbf{\textit{Fourth}}, we opt for one of the most representative RL frameworks, specifically PPO, as our backbone. Its effectiveness has been validated across various RL tasks, including renowned LLMs \cite{openai2023gpt4}. Although we perform some ablations on its actor-critic configuration, the exploration of experimental comparisons among various RL backbones, such as with model-based RLs \cite{moerland2023model}, is left for future work. 

 \textbf{Outlook}. Excitement surrounds the potential of deep learning to fully or partially replace conventional optimization in practical RT, despite acknowledged limitations. The proposed RLS stands out for three key reasons: 1) it enables potential end-to-end learning for the entire planning pipeline to seek global optimal solutions; 2) unlike the leaf sequencer method used in iterative optimizers (such as typical VMAT), there is no need to give as input a ``seed sequence'' (i.e., initial state of leaf positions and MUs) from previous iteration, allowing a more flexible algorithm design; and 3) with proper learned knowledge, RLS can potentially deliver superior performance than optimization with user-preference. We are excited about future work when more advanced algorithms and scenarios are applied.   

\newpage

\section*{Disclaimer}

The information in this paper is based on research results that are not commercially available. Future commercial availability cannot be guaranteed. 

\section*{Impact Statement}
This paper presents work whose goal is to advance the field of Machine Learning for Radiotherapy. There are many potential societal consequences of our work, none which we feel must be specifically highlighted here.

\section*{Acknowledgement}

We acknowledge the support of Varian for the research interface of practical planning environment, and its feasibility to evaluate leaf sequencing performances. 

We thank Vivek Singh and Yue Zhang from Digital Technology and Innovation, Siemens Healthineers for the helpful discussions on this study. 

We thank all the data contributors to the REQUITE project, including patients, clinicians and nurses. The core REQUITE consortium consists of David Azria, Erik Briers, Jenny Chang-Claude, Alison M. Dunning, Rebecca M. Elliott, Corinne Faivre-Finn, Sara Gutiérrez-Enríquez, Kerstie Johnson, Zoe Lingard, Tiziana Rancati, Tim Rattay, Barry S. Rosenstein, Dirk De Ruysscher, Petra Seibold, Elena Sperk, R. Paul Symonds, Hilary Stobart, Christopher Talbot, Ana Vega, Liv Veldeman, Tim Ward, Adam Webb and Catharine M.L. West. 


\bibliography{refs.bib}
\bibliographystyle{icml2024}

\newpage

\appendix
\onecolumn
\part{Appendix} 
\parttoc

Appendix \ref{background} introduces the background of radiotherapy to readers who do not have much background in radiotherapy. We include cartoon illustrations, explanations of terminologies in Appendix \ref{app:terms}, and a summary of mathematical symbols in Appendix \ref{app:symbol}.

Appendix \ref{app:PORIx} provides a more detailed introduction to the planning environment PORIx.

Appendix \ref{app:param} includes more hyperparameters compared to the main text.

Appendix \ref{app:net} details the network structures of our model.

Appendix \ref{app:othermodule} introduces the other modules (except for leaf sequencing) in the full-AI end-to-end planning pipeline, and the considerations of evaluating leaf sequencing  in 3D dose space and comparison with dose prediction metrics (Table \ref{tab:metric}).

Appendix \ref{rw_code} includes more details of reward components and their PyTorch style pseudo code.

Appendix \ref{app:data} details the datasets.

Appendix \ref{app:crop} shows multiple examples of the optimization cost curve of PORIx, comparing the proposed RLS, RLS without \textit{cropping}, and the optimization sequencer in PORIx.

Appendix \ref{app:rec} includes additional examples illustrating reconstruction errors.

\newpage
\section{Introduction to Radiotherapy}
\label{background} 

\begin{figure}[h]
    \centering
    \includegraphics[width=0.6\textwidth]{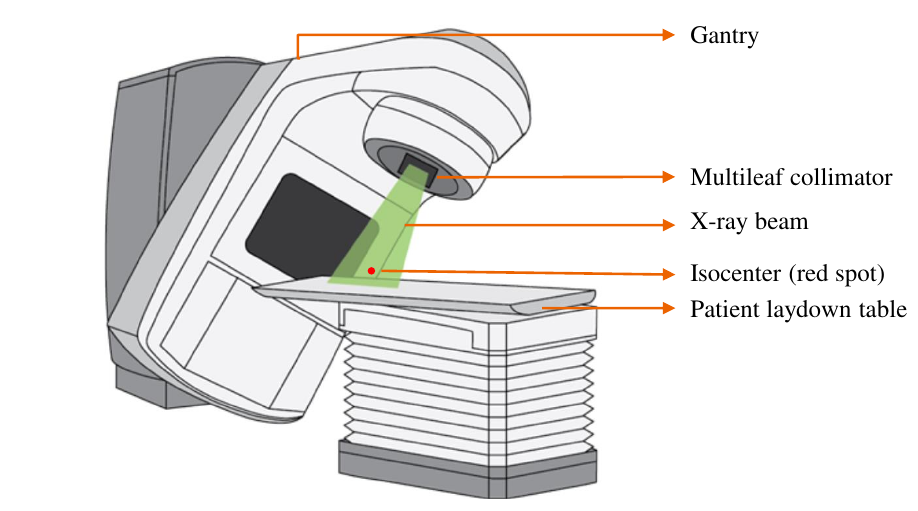}
    \caption{Cartoon illustration of a typical radiotherapy machine. The machine figure (except the annotations) is borrowed from \cite{canadianRT}. The multileaf collimator is connected with concept in Figure \ref{fig:app_mlc}.}
    \label{fig:app_rtm}
\end{figure}

\begin{figure}[h]
    \centering
    \includegraphics[width=0.9\textwidth]{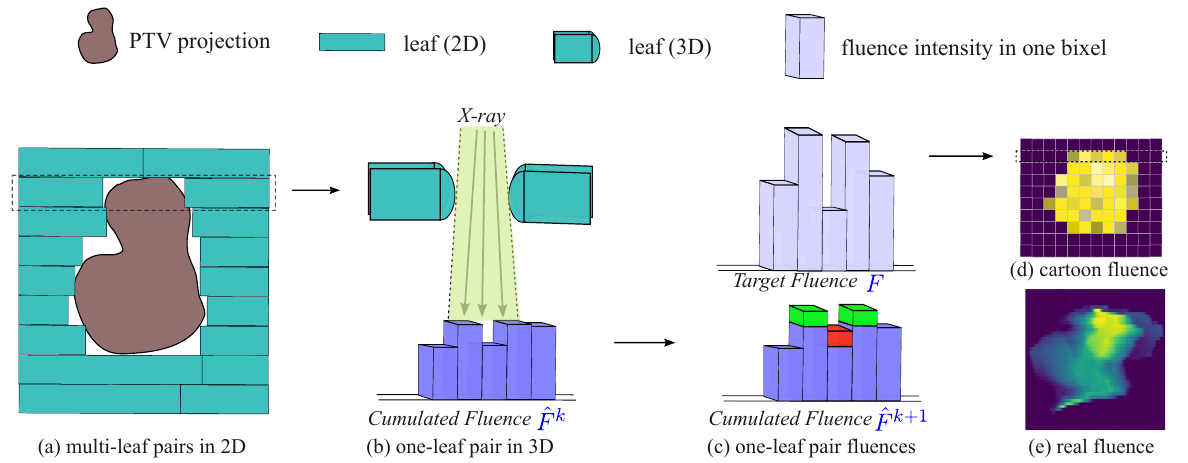}
    \caption{Cartoon illustration of MLC and fluences, an extension figure of Figure \ref{fig:mlc} in the main text. (a) shows multi-leaf pairs in 2D representing Multi-leaf Collimator and PTV projections when observed from the direction of radiation source. (b) provides a 3D view of a leaf pair and its connection to cumulated fluences. (c) illustrates motivations of Reward 1 (green) and Reward 2 (red) by comparing cumulated and target fluences. (d) shows a cartoon illustration of fluences from multi-leaf pairs, and (e) represents a real fluence map.}
    \label{fig:app_mlc}
\end{figure}

In this section, we offer a comprehensive overview of radiotherapy concepts relevant to our research. Our aim is to assist readers in understanding our work from a machine learning standpoint, even if they lack a background in radiotherapy. Explanations of related terminology and mathematical symbols are summarized in Appendix \ref{app:terms} and \ref{app:symbol}. Cartoon illustrations of the radiotherapy machine and MLC/fluences are shown in Figure \ref{fig:app_rtm} and \ref{fig:app_mlc}, respectively.

\subsection{Aberrations, Explanations of Terminologies}
\label{app:terms}
\textbf{Radiotherapy (RT):} also known as radiation therapy, it is one of the primary medical treatments that uses high-energy radiation (such as photons, electrons, or protons) to target and kill cancer cells or shrink tumors. The goal of radiation therapy is to damage the DNA inside the cancer cells, preventing them from growing and dividing, while minimizing damage to nearby healthy cells. There are two types of RT: External Beam Radiation Therapy and Internal Radiation Therapy. In this paper, we mainly focus on photon External Beam Radiation Therapy.

\textbf{Radiotherapy Planning (RTP):} 
In radiation oncology, radiotherapy planning is a precise process that strategically administers radiation to treat conditions such as cancer, with the goal of maximizing tumor destruction while minimizing damage to adjacent healthy tissues. In photon External Beam Radiation Therapy, the planning task involves optimizing incoming fields/sectors from different directions.

\textbf{Volumetric Modulated Arc Therapy (VMAT) and Intensity-Modulated Radiation Therapy (IMRT):} VMAT and IMRT are two major modern radiotherapy treatments, both aiming for precise dose delivery. In some literature, VMAT has been broadly categorized under IMRT. The illustrations of VMAT and IMRT are summarized in Figure \ref{fig:vmat_imrt}. The gantry is static for IMRT and dynamic for VMAT during X-ray delivery. 

\textbf{Planning Target Volume (PTV):} A three-dimensional (3D) volume delineated on specific medical planning images (e.g., CT images), outlining the region where the prescribed dose level is targeted to achieve the desired tumor control. The PTV generally encompasses the visually identified tumor, adjacent healthy tissue at a high risk of harboring proliferative cancer cells, and geometric margins to accommodate positioning inaccuracies between the planning images and the patient's anatomy during treatment.

\textbf{Organ at Risk (OAR):} This term refers to any healthy organ or tissue (such as the heart, lung, eye, bladder, etc.) located in proximity to the region undergoing radiation treatment, and which could potentially experience adverse effects due to exposure to radiation.

\textbf{RT Dose:} Also known as 3D Dose or Dose within the context of this paper, it denotes the precise quantity of radiation administered to a Planning Target Volume (PTV) within a patient's body throughout a course of radiation treatment. This measurement represents the absorbed amount of radiation energy by the tissues undergoing treatment. RT Dose is a 3D matrix that shares the same shape with CT and ROI contours.  

\textbf{Fluence Map:} This term refers to a depiction of the intensity of a radiation beam across the treatment field. It is a two-dimensional (2D) map that visualizes how the radiation dose is distributed across a specific cross-sectional area of the patient's body. Fluence maps play a crucial role in contemporary radiation therapy treatment planning and delivery.

\textbf{Isocenter:} The isocenter is determined during the treatment planning process, where medical professionals use imaging techniques such as CT scans to identify the tumor's location and define the treatment volume. Once the treatment plan is established, the linear accelerator or other radiation delivery equipment is adjusted to ensure that the beams intersect precisely at the isocenter. 

\textbf{Control Point:} 
This term denotes a specific time point in the administration of radiation therapy. The details of a control point encompass the positions of all leaf pairs and the monitor unit at that specific time point.

\textbf{Multileaf Collimator (MLC):} Positioned in close proximity to the patient's body and situated in the path of the radiation beam, the MLC plays a key role in shaping and controlling the radiation beam. Its primary function is to enable precise and conformal radiation treatment, facilitating highly targeted delivery.

\textbf{Leaf:} The MLC consists of a series of individual ``leaves" or ``slats" arranged in pairs. These leaves are typically made of high-density material, such as tungsten, to effectively block the radiation.

\textbf{Photon Optimizer Research Interface from Varian (PORIx):} An optimizer for photon VMAT planning that provides an interface for the necessary components for RL agent training: the implemented callbacks of the interface allows user to replace the prediction in the PORIx so the AI model can be evaluated in this environment. More details are in Section \ref{app:PORIx} and Figure \ref{fig:bluelake}. 

\textbf{Main Optimization Cost:} This term originates from the cost callbacks provided by PORIx, where it represents the dosimetric cost derived from the intermediate dose and dose objectives for each iteration in the optimization process. In Figure \ref{fig:ptv_cases} and \ref{fig:cost_crop}, the ``main optimization cost" serves as the y-axis.

\newpage
\subsection{Summary of Mathematical Symbols}
\label{app:symbol}

\textbf{Different Indices: } 

$k$: index of control points, the number of control points of a fluence map is $K$.

$x$: index of leaf agents, the number of leaf agents of a fluence map is $X$.

$y$: index of the second dimension of fluence maps, the second dimension of fluence map is $Y$, indicating the range of leaf positions is $[0, Y]$.

\textbf{Fluence Map}. The target fluence is denoted as $F$ has the size of $X\times Y$. The part of fluence associated to the $x$-th leaf pair is denoted as $F_x$. The predicted fluence is termed $\hat{F}$. The $k$-th cumulated fluence is termed as $\hat{F}^{k}$, which is cumulated from $CP_1$ to $CP_k$, the unit fluence of $CP_k$ is termed as $\tilde{F}^k$, so $\hat{F}^{k} = \sum M^k \cdot\tilde{F}^k$. Given $K$ control points for the target fluence, we have $\hat{F}$ = $\hat{F}^{K}$. 

\textbf{3D Dose}. Tn the end-to-end AI pipeline. the reference dose is termed as $D$, and predicted dose is $\hat{D}$. 

\textbf{Leaf Pair}. Tn a leaf pair, the leaf with smaller index is termed as $A$ and the other is $B$. The position of $x$-th leaf pair at $k$-th control point is termed as ($A^k_x$, $B^k_x$). 

\textbf{Control Point}: denoted as $CP$. A fluence is associated with $K$ control points, and $k$-th control point is termed as $CP_k$. 

\textbf{Monitor Unit}: termed as $M$. The monitor unit at $k$-th control point is $M^k$. 

\textbf{Leaf Policy}: termed as $\pi_{\theta_l}$. All leaf pairs share the same leaf policy. 

\textbf{MU Policy}: termed as $\pi_{\theta_m}$. 

\textbf{Critic Net}: termed as $V_{\phi}$.

\textbf{Reward}: the total reward is termed as $R$, and Reward $i$ is termed as $R_i$. As the shown in the Eq. \ref{eq:rewards}, the rewards related to $x$-th leaf pair is $R^k_i$. 

\textbf{Environment}: is termed as $Env$, which follows radiotherapy-related computations in PORIx. 

\textbf{Expection}: is termed as $\hat{\mathbb{E}}[\cdot]$.

\textbf{Sigmod Function}: is termed as $\sigma(\cdot)$. 

\textbf{Indicator Function}. In this paper, $\mathds{1}(\cdot)$ is defined as 

\[
\mathds{1}_{(0, \infty)}(x) = 
\begin{cases} 
1 & \text{if } x > 0, \\
0 & \text{if } x \leq 0.
\end{cases}
\]

\newpage
\section{Introduction of PORIx}
\label{app:PORIx}
\begin{figure}[h]
    \centering
    \includegraphics[width=0.8\textwidth]{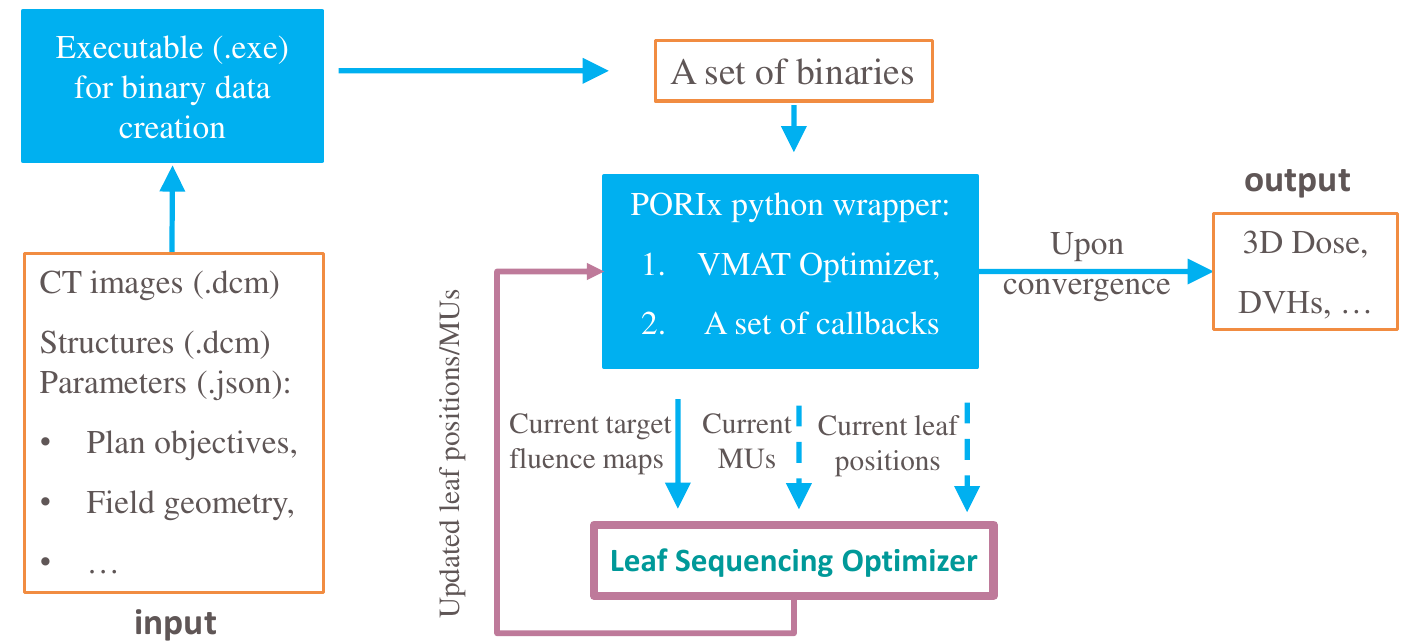}
    \caption{Illustration of RT planning with PORIx. The input of the planning includes CT, RT structures, and a set of planning parameters. The PORIx is a python wrapper of VMAT optimizer with a set of callbacks that can provide target fluences, monitor units, and leaf positions of the iterations. Leaf Sequencing Optimizer is another optimization layer for leaf sequencing only. Our RLS can replace the Leaf Sequencing Optimizer module to get an AI-integrated solution. Note our RLS do not take MUs/leaf positions from previous iteration which can work beyond a optimization environment.}
    \label{fig:bluelake}
\end{figure}

Photon Optimizer Research Interface (PORIx) originates from a prominent commercial company, Varian Medical Systems, a Siemens Healthineers company. The framework of PORIx\footnote{Our experiments are based on a version of PORIx in 2023. Some features may vary across different versions.} is depicted in Figure \ref{fig:bluelake}. This research interface is adept at generating radiotherapy (RT) plans through scripting. Callbacks within the interface allow users to substitute predictions, facilitating the evaluation of AI models within the PORIx environment. It is worth noting that clinical planning is a complex and often subjective process, involving experts with diverse backgrounds. Various metrics and considerations may be included for planning optimization, beyond the metrics used in this paper.   

In this study, our focus centers on assessing the leaf sequencing functions within the initial 100 iterations. During this phase, the number of control points in one sector is fixed at either 16 or 10, similar to the VMAT photon optimization in Varian's Eclipse. The main focus in this paper is on the first-level resolution of the multi-resolution optimization and no manual intervention during optimization. 

One potential future work could involve initiating the planning process with deep learning plans, which offer rapid and automated solutions, followed by a limited number of human/optimization intervention steps. This approach may promise a faster and more efficient path compared to conventional methods that start from scratch, while also leading to safe plans approved by experts.

We estimate the computational requirements (GPU memory occupation) of RLS and whole planning pipelines, as in Table \ref{tab:gpu_mem}. The RLS itself is computationally efficient due to the lightweight network structure described in Appendix \ref{app:net}, and it does not pose a bottleneck within end-to-end pipelines. 

\begin{table}[h]
    \centering
    \begin{tabular}{ll}
    \toprule
        Model &  Estimated GPU memory occupation\\
    \midrule
      RLS   & $\sim$0.35 GB \\
      AI E2E pipeline (dose/fluence prediction + RLS) & $\sim$2.8 GB \\
      PORIx pipeline & $\sim$3.0 GB \\
    \bottomrule
    \end{tabular}
    \caption{Estimated GPU memory occupation when running in different contexts. \textit{Disclaimer: The comparison is based on settings in this study, the number can be varied when the settings are different.}}
    \label{tab:gpu_mem}
\end{table}

\newpage
\section{Additional Settings and Hyperparameters}
\label{app:param}
\begin{Verbatim}

#--------------------------------- General Parameters ---------------------------------#

deep learning platform: PyTorch 1.13

GPU type: NVIDIA RTX A4500

RL iterations: 20000 # similar concept as args.num_iterations in CartPole(PPO) of CleanRL 

Update epochs: 2 # similar concept as args.update_epochs in CartPole(PPO) of CleanRL

batch_size: 96

#---------------------------------- Parameters for Environment -----------------------#

number of leaf agents: 50

max movement of leaf: 4

number of steps in an episode of training: 16

monitor unit range: 0.5 to 2.5

Reward 1 ratio: 1

Reward 2 ratio: 2

Reward 3 ratio: 2

Reward 4 ratio: 1

Reward 5 ratio: 1

#-------------------------------- Parameters for Loss Functions -----------------------#

discount factor gamma: 0.99

GAE lambda: 0.95

toggles advantages normalization: True

surrogate clipping coefficient: 0.2

clip value loss: True

entropy coefficient: 0.01

maximum norm for the gradient clipping: 0.5


#-------------------------------- Parameters for Optimizer ---------------------------#

learning rate: 1e-4 

weigth decay: 1e-4

optimizer: AdamW

optimizer scheduler: CosineAnnealingLR




\end{Verbatim}

\newpage
\section{Details of Network Structures}
\label{app:net}
\textit{Leaf Actor}: 
\hspace{-.2in}
\begin{Verbatim}
          ----------------------------------------------------------------
          Layer (type)           Output Shape      Param #
          ================================================================
          Input                  [-1, 205]               0
          Linear-1               [-1, 512]          105472
          ELU-1                  [-1, 512]               0
          Linear-2               [-1, 512]          262656
          ELU-2                  [-1, 512]               0
          Linear-3               [-1, 512]          262656
          ELU-3                  [-1, 512]               0
          Linear-4               [-1, 512]          262656
          ELU-4                  [-1, 512]               0
          Linear-5               [-1, 18]             9234
          ELU-5                  [-1, 18]                0
          ================================================================
          Total params: 902674 
          Trainable params: 902674   
          ----------------------------------------------------------------
\end{Verbatim}
\textit{MU Actor}: 
\hspace{-.2in}
\begin{Verbatim}
          ----------------------------------------------------------------
          Layer (type)           Output Shape      Param #
          ================================================================
          Input                  [-1, 205]               0
          Linear-1               [-1, 512]          105472
          ELU-1                  [-1, 512]               0
          Linear-2               [-1, 512]          262656
          ELU-2                  [-1, 512]               0
          Linear-3               [-1, 512]          262656
          ELU-3                  [-1, 512]               0
          Linear-4               [-1, 2]              1026 
          ELU-4                  [-1, 2]                 0
          ================================================================
          Total params: 631810 
          Trainable params: 631810   
          ----------------------------------------------------------------
\end{Verbatim}
\textit{Critic}:
\hspace{-.02in}
\begin{Verbatim}
          ----------------------------------------------------------------
          Layer (type)           Output Shape      Param #
          ================================================================
          Input                  [-1, 205]               0
          Linear-1               [-1, 512]          105472
          ELU-1                  [-1, 512]               0
          Linear-2               [-1, 512]          262656
          ELU-2                  [-1, 512]               0
          Linear-3               [-1, 512]          262656
          ELU-3                  [-1, 512]               0
          Linear-4               [-1, 512]          262656
          ELU-4                  [-1, 512]               0
          Linear-5               [-1, 1]               513
          ELU-5                  [-1, 1]                 0
          ================================================================
          Total params: 893953 
          Trainable params: 893953   
          ----------------------------------------------------------------
\end{Verbatim}

\newpage
\section{The Full-AI End-to-end (E2E) Pipeline}
\label{app:othermodule}

\subsection{Introduction of Other Modules}

\textbf{Dose Prediction}.
Dose prediction with deep learning is a relatively well-studied topic. It can be considered an extension of planning objectives prediction compared to the dot objectives in conventional optimization or line objectives in RapidPlan. In most literature, the input of this module includes the CT and masks of PTV/OARs (along with other auxiliary conditions), and the output is the estimated 3D dose required for cancer treatment. 

In this work, the dose prediction module is trained in a conditional GAN framework, inspired by \citet{gao2023flexible,Kearney2020DoseGAN:Generation}. The adversarial training is to make the predicted 3D dose realistic. In the end-to-end AI pipeline, this dose prediction module serves as the \textit{dose objectives definition} as in Figure \ref{fig:PTP}. 

\textbf{Fluence Prediction}. After obtaining the 3D dose, we predict this fluence map of each beam angle as a 2D intensity map. This module is motivated by \citet{Wang2020FluenceTherapy} and we have made significant changes to extend the fluence map prediction for VMAT. The fluence prediction module predicts all fluence maps jointly, whose input is projections of the 3D dose map to the beam's eye view of gantry angles. In the end-to-end AI pipeline, this dose prediction module serves as the \textit{optimal fluence prediction} as in Figure \ref{fig:PTP}. Following a similar setting in PORIx that multiple control points contribute to each fluence, the resolution of target fluences in the VMAT context is eight control points per fluence except the beginning and ending fluences is nine control points (as in bottom of Table \ref{tab:metric}). A full arc has 178 control points in total, which has been divided into 22 sectors for leaf sequencing. 

The primary difference between this full-AI pipeline and frequently used optimization methods lies in its ability to be executed in a single (or few) iteration(s), thus the former can potentially accelerate the planning process significantly. It is important to note that this full-AI pipeline is currently in the preliminary stage and may yield lower quality plans than commercial optimization platforms. Also, some detailed clinical settings may be simplified in this AI pipeline. Despite these considerations, we see the full-AI pipeline as a valuable testbed, particularly since this study mainly focuses on evaluating the leaf sequencing module.

\subsection{Consideration of Table \ref{tab:metric} }

The rationale behind utilizing the dose prediction record presented in Table \ref{tab:metric} is to assess the error scale of our model by comparing it to an extensively studied task (i.e., dose prediction) in a full-AI end-to-end pipeline. A good match between the target and prediction indicates promising performance of RLS in this end-to-end pipeline. Notably, there is currently no established reference in the literature for a direct comparison with our model in such a pipeline. The decision to refer to the OpenKBP record is based on its status as the most widely recognized open challenge in radiotherapy, with researchers consistently updating and improving the top-performing models through open-sourced evaluations. We acknowledge those two (i.e., dose prediction (upper part of Table \ref{tab:metric}) and evaluation fluence prediction in 3D dose perspective (lower part of Table \ref{tab:metric}) \footnote{The number of test patients for end-to-end pipeline is 64 since treatment configurations of some patients are not applicable to dose/fluence prediction modules.} are not the same task and show their similarity and differences below: 

\textbf{Similarity}: both those two tasks are evaluated by the difference between two 3D dose maps based modules in radiotherapy plan workflow. 

\textbf{Differences}: as mentioned above, the outcome in dose prediction is a 3D estimation of dose distribution, and there is no guarantee that the machine can achieve the predicted dose. Contrastively, the two doses from the second task are obtained by the same dose calculation method (i.e., Acuros XB of Eclipse\footnote{\url{https://www.varian.com/products/radiotherapy/treatment-planning/eclipse}}) and the target/prediction fluences. Fundamentally, this is for evaluating the performance of leaf sequencing but using an auxiliary way. The predicted dose of this module is machine executable.

\newpage
\section{Details/Pseudo-Code of Rewards}

This section provides supplementary information for rewards in Section \ref{sec:definition}. The trade-off between reconstruction accuracy and other clinically relevant factors is carefully considered in the design of rewards. Specifically, Reward 1 and 2 are to guide agents to reconstruct the target fluence, and Reward 3, 4, and 5 take into account various clinical considerations.

\textbf{Reward 1}. As depicted in Figure \ref{fig:app_mlc}c, the green parts indicate the Reward 1, which pushes the cumulative fluence closer to target fluence. 

\textbf{Reward 2}. As depicted in Figure \ref{fig:app_mlc}c, the red parts indicate Reward 2, which punishes agents if the intensity of cumulative fluence exceeds that of target fluence. 

\textbf{Reward 3}: avoiding cross leaves. This reward component is to encourage the agent to avoid predictions where left- and right-leaf positions intersect, which violates the physical laws. In addition to this, the environment enforces that the final leaf positions do not intersect and do not extend beyond the radiation fields.

\textbf{Reward 4}: smoothness of actions (Leaf/MU change between control-points). Ensuring smooth change in leaf / MU configuration is necessary as significant or sudden changes in leaf positions may cause machine instability. We demonstrated that by tuning the weight of the reward (Figure 11), our model allows users to balance between the maximum leaf speed (machine-dependent parameter) and the reconstruction performance.

\textbf{Reward 5}: aperture shape regularization motivated from \citet{younge2012penalization}. According to Younge et al., volumetric modulated arc therapy (VMAT) planning often yields small, irregular aperture shapes, leading to dosimetric inaccuracies during delivery. We included this regularization as R5 = area of aperture / perimeter of aperture following \citet{younge2012penalization}.

Below shows the Pytorch-style pseudo code of the rewards:

\label{rw_code}
\begin{lstlisting}[language=Python] %, caption=Python example

# MU: monitor unit,  mask: unit fluence
# y_tail: leaf position of leaf A_x (i.e., the left leaf or the tail leaf)
# y_front: leaf position of leaf B_x (i.e., the right leaf or the front leaf)
rw1 = MU * torch.sum((tar_fluence - cumu_fluence >= MU) * mask, axis = 1)
cumu_fluence += mask * MU
rw2 = - torch.sum((tar_fluence - cumu_fluence < 0) * mask, axis=1)
rw3 = (y_front - y_tail > 0) - (y_front - y_tail < 0)
rw4 = 3 - sigmoid(delta_front) - sigmoid(delta_tail) - sigmoid(delta_MU) 
rw5 = Aperture(y_front, y_tail)
\end{lstlisting}

\newpage

\section{Dataset Description}
\label{app:data}
The dataset distribution is described in  Table \ref{tab:data_dist}. Unlike tasks in other AI fields (e.g., computer vision or natural language processing tasks are usually easy to get thousands, millions, or even billions of data samples), RT planning data for research use is relatively limited. Most RTP research related to deep learning typically involves fewer than 500 patients. For example, the well-studied OpenKBP challenge includes only 340 patients. Our study is conducted on relatively larger-scale data resources with more patients. 
\begin{table}[h]
    \centering
    \begin{tabular}{cccccc}
    \toprule
          Dataset Name & Cancer Site & \# Patient after filtering &  train/val/test & Availability \\
          \midrule
          \vspace{.2in}
        HNd & head-and-neck & 493 & 370/10/113 & \parbox{5cm}{Data will not be public available due to data regularization} \\
        \vspace{.2in}
        HNe1 & head-and-neck & 31 & 0/0/31 & \parbox{6cm}{Public: www.cancerimagingarchive\\.net/collection/hnscc-3dct-rt} \\
        \vspace{.2in}
        HNe2 & head-and-neck & 140 & 0/0/140 & \parbox{6cm}{Public: www.cancerimagingarchive\\.net/collection/hnscc} \\
        \vspace{.2in}
        Pros & Prostate & 555 & 471/10/84 &\parbox{6cm}{Public with permission required \cite{Seibold2019REQUITE:Cancer}} \\ 
        Pros(e) & \multicolumn{4}{c}{the same set as Pros, but test results are reported when model trained by head-and-neck site.}\\
        \bottomrule
    \end{tabular}
    \vspace{-0.1in}
    \caption{The description of datasets used in our experiments. Note that our experiment is conducted at fluence level. Each patient has up to 525 target fluences (within 100 iterations optimization in PORIx). Compared to original dataset, some patients are filtered due to missing data, quality issue, and incompatibility with environment/software.}
    \label{tab:data_dist}
\end{table}

The experiments and evaluations on different datasets/sites/scenarios (e.g., Table \ref{tab:rec_err} and \ref{tab:iter_num}) indicate the generalizability of our model. 
Furthermore, leaf sequencing is executed at the fluence-level, encompassing a sample size that is hundreds of times larger than the number of patients. Our model can have a reasonably good performance even when the training size is smaller, as in Table \ref{tab:num_of_train}. 

\begin{table}[h]
    \centering
    \begin{tabular}{ll}
    \toprule
        Model &  MNSE ($\downarrow$) \\
    \midrule
      PORIx   & 0.219 \\
      RLS (trained with 50 patients) & 0.158 \\
      RLS (trained with 100 patients) & 0.152 \\
      RLS (trained with 370 patients) & 0.149 \\
    \bottomrule
    \end{tabular}
    \caption{The reconstruction error MNSE of different number of training patients.}
    \label{tab:num_of_train}
\end{table}

\section{Supplementary Examples of \textit{Cropping} Ablations}
\label{app:crop}
\begin{figure}
    \centering
    \includegraphics[width=0.95\textwidth]{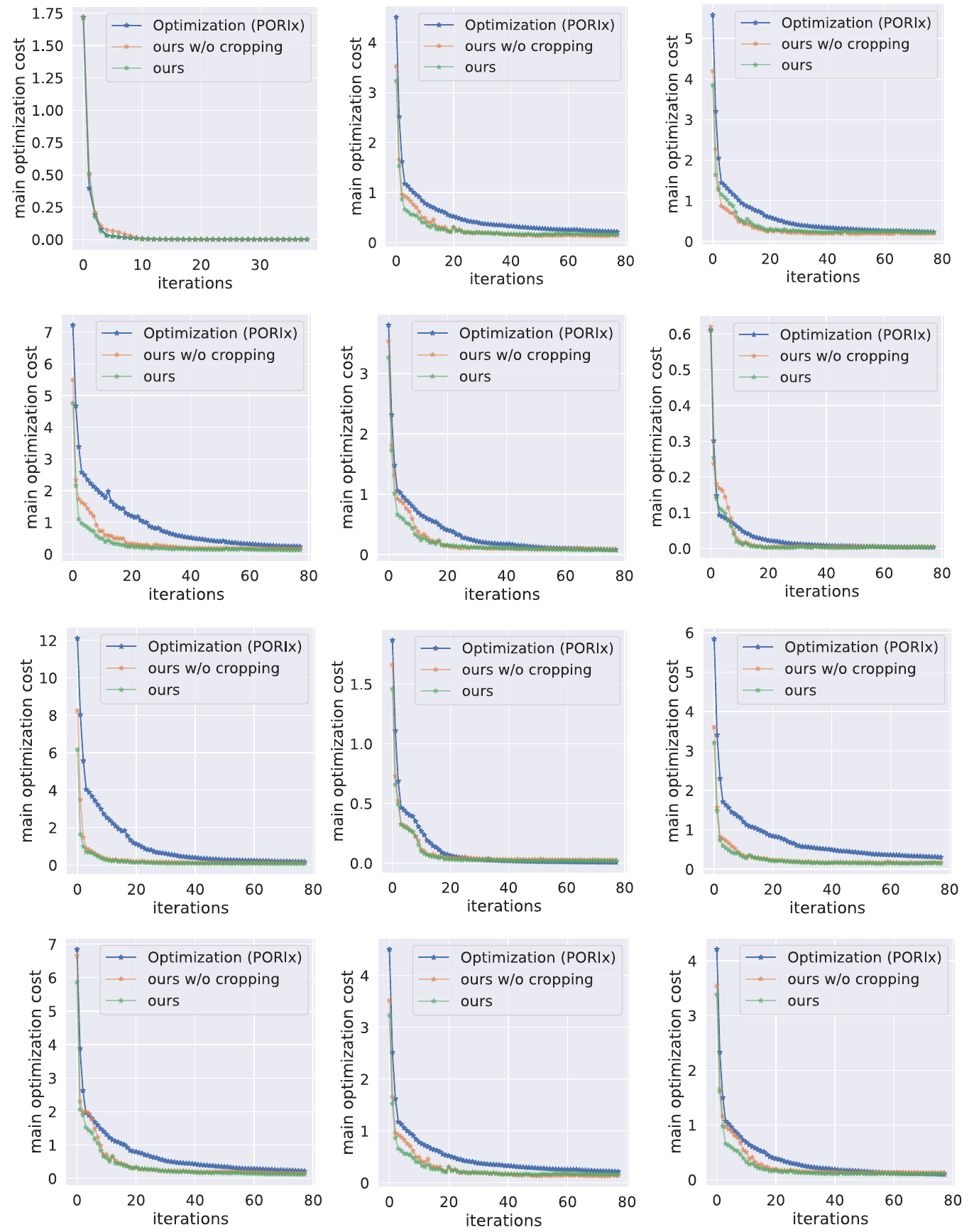}
    \caption{Ablation study with vs. without \textit{cropping}. The main optimization cost is from callback of PORIx planning environment. \textbf{ours} denotes that our RLS has replaced the leaf sequencing optimizer of PORIx. }
    \label{fig:cost_crop}
\end{figure}

As shown in Figure \ref{fig:cost_crop}, the \textit{cropping} strategy helped the optimization in terms of faster convergence. Overall, our RLS can achieve smaller error and faster convergence especially in early iterations during the optimization.   

\section{Additional Examples Illustrating Reconstruction Errors}
\label{app:rec}
\begin{figure}[h]
    \centering
    \includegraphics[width=0.98\textwidth]{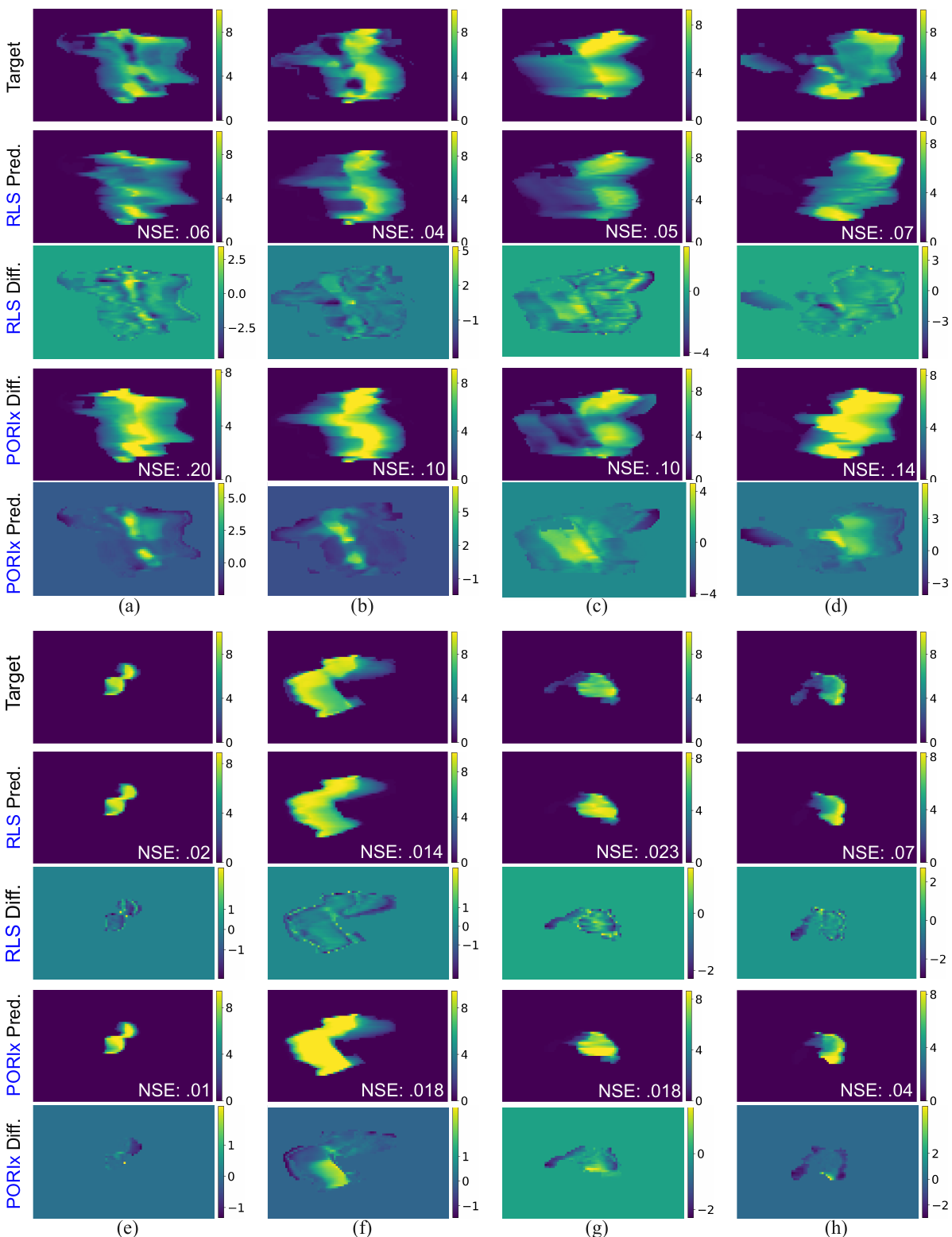}
    \caption{Supplement examples of Figure \ref{fig:example_fluence} in the main text. We provide difference maps between predicted and target fluences here.}
    \label{fig:app_rec_flu}
\end{figure}

Continued with Figure \ref{fig:example_fluence}, we supplement more and detailed illustrations of fluence reconstruction in Figure \ref{fig:app_rec_flu}. Figure \ref{fig:app_rec_flu}a to Figure \ref{fig:app_rec_flu}h represents examples from eight different patients, where (a) - (f) are head-and-neck cancer cases, and (g) - (h) are from prostate cancer site. For each case, the five figures represent \textbf{target fluence}, \textbf{predicted fluence from RLS}, \textbf{difference between RLS prediction and target}, \textbf{predicted fluence from PORIx}, \textbf{difference between PORIx prediction and target}, from up to down. The reconstruction error is shown at right bottom. RLS achieves smaller error in the majority of cases. 
\end{document}